\documentclass{article} 
\usepackage[final]{colm2026_conference}

\PassOptionsToPackage{numbers, sort&compress}{natbib}

\usepackage{booktabs}
\usepackage{colortbl}
\usepackage[dvipsnames]{xcolor}
\usepackage{graphicx}
\usepackage{array}
\usepackage{tabularx}
\usepackage[utf8]{inputenc}
\usepackage[T1]{fontenc}
\usepackage{hyperref}
\usepackage{url}
\usepackage{booktabs}
\usepackage{amsfonts}
\usepackage{nicefrac}
\usepackage{microtype}
\usepackage[utf8]{inputenc}
\usepackage{booktabs}
\usepackage{tabularx}
\usepackage{caption}
\usepackage{array}
\usepackage{tabularx}
\usepackage{makecell}
\usepackage{siunitx}
\usepackage{amsmath}
\usepackage{soul}
\usepackage{array}
\usepackage{courier}
\usepackage{verbatim}
\usepackage{framed}
\usepackage{tcolorbox}
\usepackage{enumitem}
\usepackage{wrapfig}
\tcbuselibrary{listingsutf8}
\usepackage{caption}
\DeclareCaptionType{prompt}[Prompt][List of Prompts]
\DeclareCaptionType{database}[Database Tables][List of Database Tables]
\definecolor{highlightrow}{HTML}{FFF2CC}
\usepackage{pifont}
\newcommand{\cmark}{\ding{51}}%
\newcommand{\xmark}{\ding{55}}%
\usepackage{fancyhdr}
\usepackage{float}
\usepackage{tabularx}
\usepackage{booktabs}
\usepackage{lineno}
\usepackage{caption} 
\newfloat{database}{tbp}{lod}
\floatname{database}{Database Table}

\definecolor{darkblue}{rgb}{0, 0, 0.5}
\hypersetup{colorlinks=true, citecolor=darkblue, linkcolor=darkblue, urlcolor=darkblue}

\newtcblisting{promptbox}{
  colback   = white!5,
  colframe  = black!70,
  listing only,
  listing engine = listings,
  listing options = {
      basicstyle=\ttfamily\tiny,
      escapeinside={(*@}{@*)},
  },
  left=1pt, right=1pt, top=0pt, bottom=0pt,
  boxrule=0.4pt, arc=2pt
}

\newcommand{\pmval}[1]{{\scriptsize{(±#1)}}}

\newcommand{\name}{BiomedSQL}
\newcommand{\farazrewrote}[1]{}
\title{\name: Text-to-SQL for Scientific Reasoning on \\ Biomedical Knowledge Bases}

\author{
  Mathew J. Koretsky\textsuperscript{\textnormal{1,2}\thanks{Corresponding author email: mathew@datatecnica.com}},
  Maya Willey\textsuperscript{\textnormal{1,2}},
  Owen Bianchi\textsuperscript{\textnormal{1,2}},
  Chelsea X. Alvarado\textsuperscript{\textnormal{1,2}}, \\
  \textbf{Tanay Nayak}\textsuperscript{\textnormal{2,3}},
  \textbf{Nicole Kuznetsov}\textsuperscript{\textnormal{1,2}},
  \textbf{Sungwon Kim}\textsuperscript{\textnormal{2,3}},
  \textbf{Mike A. Nalls}\textsuperscript{\textnormal{1,2,4}}, \\
  \textbf{Daniel Khashabi}\textsuperscript{\textnormal{2,3}},
  \textbf{Faraz Faghri}\textsuperscript{1,2,4} \\[1ex]
  \textsuperscript{1}Center for Alzheimer’s and Related Dementias, NIA, NIH; 
  \textsuperscript{2}DataTecnica LLC; \\
  \textsuperscript{3}Johns Hopkins University; 
  \textsuperscript{4}Laboratory of Neurogenetics, NIA, NIH
}

\begin{document}

\ifcolmsubmission
\linenumbers
\fi

\maketitle

\begin{abstract}
Biomedical researchers increasingly rely on large-scale structured databases for complex analytical tasks. However, current text-to-SQL systems often struggle to map qualitative scientific questions into executable SQL, particularly when implicit domain reasoning is required. We introduce \emph{\name}, the first benchmark explicitly designed to evaluate scientific reasoning in text-to-SQL generation over a real-world biomedical knowledge base. \name{} comprises 68,000 question/SQL query/answer triples generated from templates and grounded in a harmonized BigQuery database that integrates gene–disease associations, causal inference from omics data, and drug approval records. Each question requires models to infer domain-specific criteria, such as genome-wide significance thresholds, effect directionality, or trial phase filtering, rather than rely on syntactic translation alone. We evaluate a range of open- and closed-source LLMs across prompting strategies and interaction paradigms. Our results reveal a substantial performance gap: Gemini-3-Pro achieves 58.1\% execution accuracy under baseline prompting, while our custom multi-step agent, BMSQL, reaches 62.6\%, both well below the expert baseline of 90.0\%. \name{} provides a new foundation for advancing text-to-SQL systems that support scientific discovery through robust reasoning over structured biomedical knowledge bases. The BiomedSQL benchmark and codebase are publicly available at \url{https://datatecnica.github.io/biomedbench-suite/biomedsql}.
\end{abstract}

\section{Introduction}
\label{sec:intro}
Modern biomedical research is increasingly data-centric. Electronic health records (EHRs), high-throughput assays, and population-scale studies populate large databases that researchers query daily. Natural-language interfaces, particularly text-to-SQL systems, offer the promise of democratizing access to these resources. However, most current systems treat query generation as a syntactic translation task, mapping question structure to SQL syntax without deeper domain understanding. This abstraction breaks down in biomedical contexts. Domain experts routinely ask questions such as \textit{“What SNPs are most significantly associated with Alzheimer’s disease?”} or \textit{“What drugs target genes up-regulated in Parkinson’s disease?”}---questions grounded in scientific conventions, such as statistical thresholds, drug approval pathways, and causal inference across modalities.

\begin{figure}[htbp]
    \centering
    \includegraphics[width=1.0\textwidth]{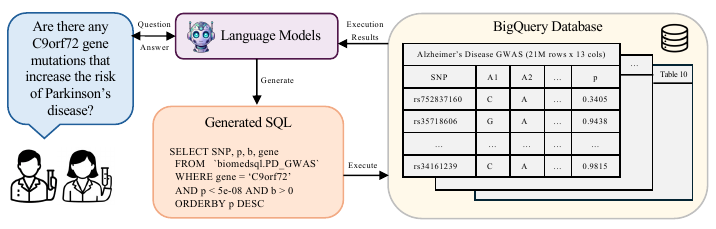}
    \caption{Example text-to-SQL workflow used to evaluate LLM performance on BiomedSQL. Given a question and the database schema, an LLM must generate a SQL query and use its execution results to return a natural language response.}
    \label{fig:workflow}
\end{figure}

These domain-specific conventions---e.g., genome-wide significance cutoffs, trial phase filtering, or effect-size interpretation---are invisible from the schema alone. While general-purpose text-to-SQL benchmarks (e.g., \textsc{Spider}~\citep{spider2018}, BIRD~\citep{Li2023BIRD}) have advanced the field, they do not evaluate the scientific reasoning required in complex domains. Similarly, EHR-focused benchmarks~\citep{Wang2020MIMICSQL, lee2022ehrsql} emphasize temporal logic or patient retrieval, but do not isolate or rigorously test \textbf{scientific reasoning on large-scale databases} that is required for interpreting biomedical data. This includes inferring statistical significance thresholds or chaining multi-step filtering logic across ontologies the way a biomedical analyst would.

To address this gap, we introduce \textbf{\name}, the first benchmark specifically designed to evaluate \textbf{scientific reasoning in SQL generation} within the biomedical domain. \name{} contains 68,000 biomedical question/SQL query/answer triples that reflect realistic, complex scientific queries. These queries were templated from a set of 40 unique questions that were generated and verified by domain experts for their real-world biomedical research impact and ability to be answered through the use of structured data. These are executable against a harmonized, publicly available BigQuery database integrating gene–disease associations, multi‑omic causal inferences, and drug approval records. These data sources were also verified as being the most up-to-date by domain experts that use them in their day-to-day computational research activities. Each question in \name{} requires models to translate nuanced, qualitative scientific language into executable SQL logic. 

Figure \ref{fig:workflow} provides an overview of our evaluation pipeline, where we task various LLMs in a variety of prompting and interaction scenarios on their ability to generate accurate SQL queries and summarize the execution results into a concise, natural language response based on the provided question. Our evaluation on BiomedSQL indicates significant room for improvement in deploying LLMs on biomedical text-to-SQL tasks. The top-performing model, Gemini-3-Pro, only achieves an execution accuracy of 58.1\%. Our custom multi-step agent, BMSQL, improves this to 62.6\%, but still trails expert performance (90\%).

\textbf{Our contributions are as follows:} (a) \textbf{Benchmark:} We introduce \name{}, a benchmark of 68,000 augmented question/SQL query/answer triples designed to evaluate scientific reasoning capabilities in text‑to‑SQL systems on a realistic biomedical database. (b) \textbf{Infrastructure:} We release a BigQuery database, expert-authored gold queries, execution results, and a toolkit for evaluation. (c) \textbf{Evaluation:} We assess a range of models, prompting styles, and interaction paradigms, including a custom-built text-to-SQL system (BMSQL), revealing a 25-30 percentage point gap compared to domain expert performance.


\section{Related Work}
We organize related work into four strands: (1) general-purpose text-to-SQL benchmarks, (2) text-to-SQL benchmarks for clinical databases, (3) evaluations of scientific reasoning in NLP, and (4) a direct comparison between BiomedSQL and previous benchmarks.
 
\textbf{General-purpose text-to-SQL benchmarks.} Text-to-SQL research has been largely driven by cross-domain benchmarks. Early work such as Seq2SQL~\citep{zhong2017seq2sql} introduced SQL generation for simple, single-table queries. \textsc{Spider}~\citep{spider2018} expanded generalization challenges by spanning 200 multi-table databases, catalyzing the development of schema linking techniques. More recently, KaggleDBQA~\citep{lee2021kaggledbqa} and BIRD~\citep{Li2023BIRD} added realism by incorporating enterprise-scale data and requiring attention to data quality, joins, and latency. Despite progress, top models still underperform humans by 15-20\% on execution accuracy for general-purpose benchmarks like BIRD~\citep{Li2023BIRD}. These benchmarks emphasize syntactic translation and schema generalization, not the reasoning that underlies many queries in scientific disciplines. In contrast, \name{} uses a multi-table schema enriched with domain-specific semantics, requiring models to interpret statistical conventions and biomedical ontologies---capabilities not previously assessed.

\textbf{Text-to-SQL benchmarks for clinical databases.} Several efforts have targeted clinical database querying for EHRs. \textsc{MIMICSQL}~\citep{Wang2020MIMICSQL} presented a synthetic SQL benchmark over MIMIC-III but was limited by its narrow schema. \textsc{EHRSQL}~\citep{lee2022ehrsql} advanced realism by crowdsourcing 1,000+ natural language queries from clinicians across MIMIC-III and eICU, highlighting challenges in temporal logic and data sparsity. Recent datasets have diversified queries across relational, document, and graph data models~\citep{sivasubramaniam2024sm3}, reflecting the increasing heterogeneity of EHR data. These benchmarks focus on patient-centric information retrieval and assess the ability to map schema-dependent queries. Conversely, \name{} targets discovery in biomedical research, where queries require implicit scientific reasoning such as applying significance thresholds, combining multi-omic associations, or resolving drug–gene–disease relationships, offering a complementary evaluation focused on reasoning-heavy data exploration.

\textbf{Evaluating scientific reasoning in NLP.}
Scientific reasoning has emerged as a critical frontier in NLP for tasks requiring multi-hop inference, evidence synthesis, and structured decision-making. Benchmarks such as SciFact~\citep{wadden2020scifact} and EntailmentBank~\citep{dalvi2021explaining} evaluate scientific claim verification and multi-step reasoning over textual evidence. Prompting techniques like Chain-of-Thought~\citep{wei2022chain} and ReAct~\citep{yao2023react} have demonstrated improved performance on multi-step reasoning tasks in both general and biomedical settings. More recent efforts have extended evaluation to structured data, such as SQL-R1~\citep{ma2025sqlr1} and Science Hierarchography~\citep{gao2025sciencehierarchy}, which assess LLM reasoning over ontologies and relational programs. Despite these advances, critical challenges remain in aligning model reasoning with biomedical standards of rigor, safety, and explainability. \name{} addresses this by evaluating models on their ability to infer and operationalize scientific reasoning, including statistical thresholds, ontology resolution, and complex filtering, in text-to-SQL generation over large biomedical knowledge bases.

\textbf{Benchmark comparison.}  
We compare \name{} to several recent text-to-SQL benchmarks in Table~\ref{tab:sql_benchmarks}. BiomedSQL is unique in four ways:
(1) it contains the largest number of question/SQL query/answer triples,  
(2) it features some of the longest average SQL queries (second to EHRSQL),  
(3) it explicitly targets scientific reasoning rather than schema translation, and  
(4) it evaluates both the generated SQL and the model's natural language response.
BiomedSQL is also the only biomedical domain-specific benchmark that tests reasoning beyond the scope of the schema itself, and the only one using BigQuery, a cloud-native SQL dialect, thus further simulating deployment environments.

\begin{table}[htbp]
    \setlength{\tabcolsep}{3pt}
    \centering
    \small
    \caption{Comparison of \name{} to other text-to-SQL benchmarks. \name{} uniquely evaluates scientific reasoning, natural language responses, and BigQuery SQL generation.}
    \label{tab:sql_benchmarks}
    \resizebox{\textwidth}{!}{
    \begin{tabular}{cccccccccc}
        \toprule
        \textbf{Dataset} & \textbf{\shortstack{Number of\\ Questions}} & \textbf{\shortstack{Number of\\ Queries}} & \textbf{\shortstack{Avg.\\Tokens}} & \textbf{Knowledge} & \textbf{Template} & \textbf{\shortstack{Scientific\\Reasoning}} & \textbf{\shortstack{NL\\Response}} & \textbf{BigQuery} \\
        \midrule
        MIMICSQL~\citep{Wang2020MIMICSQL} & 10,000 & 10,000 & 57.4 & \textcolor{Red}{\xmark} & \textcolor{Green}{\cmark} & \textcolor{Red}{\xmark} & \textcolor{Red}{\xmark} & \textcolor{Red}{\xmark} \\
        EHRSQL~\citep{lee2022ehrsql} & 24,000 & 24,000 & \textbf{109.9} & \textcolor{Red}{\xmark} & \textcolor{Green}{\cmark} & \textcolor{Red}{\xmark} & \textcolor{Red}{\xmark} & \textcolor{Red}{\xmark} \\
        SM3~\citep{sivasubramaniam2024sm3} & 10,000 & 40,000 & 26.1 & \textcolor{Red}{\xmark} & \textcolor{Green}{\cmark} & \textcolor{Red}{\xmark} & \textcolor{Red}{\xmark} & \textcolor{Red}{\xmark} \\
        BIRD~\citep{Li2023BIRD} & 12,751 & 12,751 & 50.6 & \textcolor{Green}{\cmark} & \textcolor{Red}{\xmark} & \textcolor{Red}{\xmark} & \textcolor{Red}{\xmark} & \textcolor{Red}{\xmark} \\
        \midrule
        \textbf{BiomedSQL} & \textbf{68,227} & \textbf{68,227} & 96.4 & \textcolor{Green}{\cmark} & \textcolor{Green}{\cmark} & \textcolor{Green}{\cmark} & \textcolor{Green}{\cmark} & \textcolor{Green}{\cmark} \\
        \bottomrule
    \end{tabular}
    }
\end{table}

\section{BiomedSQL Construction}
\label{sec:biomedsql}
BiomedSQL is designed to evaluate scientific reasoning in text-to-SQL generation over structured biomedical knowledge. We construct it by: (1) harmonizing a multi-source relational database to support biomedical queries, (2) authoring gold-standard SQL annotations from domain experts, and (3) augmenting the dataset from templates to produce 68,000 question/SQL query/answer triples.

\subsection{Relational Database Construction}
\label{subsec:1}
We constructed a relational database spanning ten core tables drawn from trusted biomedical resources, ensuring sufficient coverage for answering the full set of 68,000 questions. Data was preprocessed for consistency and deployed to BigQuery.

Our primary data sources include the \textbf{Open Targets Platform}~\citep{opentargets}, which aggregates gene–disease–drug associations, and \textbf{ChEMBL}~\citep{CHEMBL}, a curated database of bioactive molecules and pharmacological data. Open Targets data was retrieved via FTP, while ChEMBL data was accessed through BigQuery. These sources provide a unified schema of gene–disease links, drug–target pairs, trial status, and pharmacodynamics.

To support questions involving statistical genetics, we included summary statistics from large-scale \textbf{GWAS studies of Alzheimer’s disease}~\citep{bellenguez2022new} \textbf{and Parkinson’s disease}~\citep{nalls2019parkinsons}, obtained from the GWAS Catalog. We retained SNP-level data including p-values, rsIDs, allele frequencies, and nearest-gene mappings.

We integrated causal inference data from \textbf{omicSynth}~\citep{alvarado2024omicsynth}, which applies summary-data-based Mendelian randomization (SMR) to identify multiomic biomarkers with putative causal links to neurodegenerative diseases. These datasets enable reasoning over associations not directly stated but statistically inferred, e.g., \textit{“What metabolites causally influence Parkinson’s progression?”}.

All tables were normalized and uploaded to Google Cloud BigQuery. A full schema is provided in Appendix~\ref{subsec:tables}. To support future expansions, we have curated additional tables that extend BiomedSQL’s coverage to broader omics and clinical-trial data.

\subsection{SQL Annotation and Augmentation}
\label{subsec:2}
\textbf{Gold SQL authoring.}  
To ensure executable grounding, a domain expert manually wrote gold-standard SQL queries for each of the 40 seed questions drawn from CARDBiomedBench~\citep{bianchi2026cardbiomedbench}. Each query retrieves the minimum evidence necessary to answer the corresponding question, avoiding \verb|SELECT *| patterns and capping results at 100 rows. Two additional biomedical analysts independently reviewed every query, its execution output, and the corresponding natural language answer for syntactic correctness and semantic fidelity. Disagreements were resolved through iterative discussion and revision to produce a single consensus annotation. Because separate annotations were not retained prior to adjudication, formal inter-annotator agreement statistics were not computed.

\textbf{Programmatic scaling.}  
Each of the 40 queries was then templated and expanded by programmatically inserting disease, gene, and SNP mentions, aligning them to the full set of 68,000 QA pairs in CARDBiomedBench. All templated SQL queries were then executed on BigQuery to obtain ground-truth results for evaluating LLM-generated queries. This pipeline produced a benchmark where each QA pair is linked to an executable SQL query and its execution result, enabling precise evaluation of models’ ability to translate scientific questions into domain-grounded, semantically valid, and executable SQL logic.

\section{Dataset Analysis}
\label{sec:dataset_analysis}

\begin{figure}[t]
\centering

\hfill
\begin{minipage}[t]{0.61\textwidth}
    \vspace{0pt}
    \centering
    \footnotesize
    \setlength{\tabcolsep}{1pt}
    \renewcommand{\arraystretch}{0.95}

    \captionof{table}{Description of SQL categories in BiomedSQL.}
    \label{tab:sql_categories}

    \begin{tabularx}{\linewidth}{
        @{}
        >{\raggedright\arraybackslash}p{0.28\linewidth}
        @{\hspace{0.3em}}
        >{\raggedright\arraybackslash}X
        @{}
    }
        \toprule
        \textbf{SQL Category} & \textbf{Description} \\
        \midrule
        Select &
        Retrieves columns from one or more tables. \\
        Distinct &
        Retrieves unique values from columns. \\
        Join &
        Combines rows across multiple tables. \\
        Multi-Filter &
        Applies compound filters (AND, OR, NOT). \\
        Threshold &
        Filters using logical or statistical thresholds. \\
        Calculate &
        Performs arithmetic operations. \\
        Order By &
        Sorts the result set by specified columns. \\
        Similarity Search &
        Performs pattern-based retrieval. \\
        \bottomrule
    \end{tabularx}
\end{minipage}
\hfill
\begin{minipage}[t]{0.35\textwidth}
    \vspace{0pt}
    \centering
    \includegraphics[width=\linewidth]{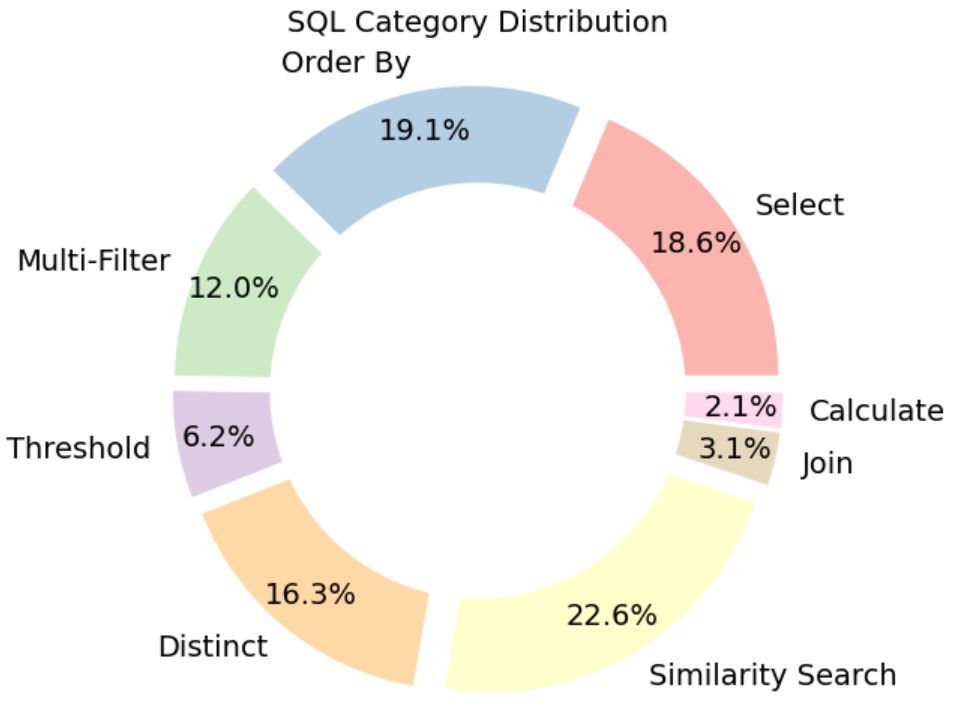}
    \captionof{figure}{SQL categories.}
    \label{fig:dataset_distributions}
\end{minipage}

\end{figure}

\textbf{Task distribution and SQL complexity.}  
To characterize the scientific and computational complexity of \name{}, we annotated all 68,000 question–query pairs with SQL operation types and biomedical reasoning categories. Table~\ref{tab:sql_categories} defines the SQL operation categories, and Figure~\ref{fig:dataset_distributions} shows their empirical distribution. Simpler operations—such as \emph{Select}, \emph{Order By}, and \emph{Calculate}—require relatively shallow syntactic parsing, which LLMs tend to perform well on. In contrast, operations such as \emph{Multi-Filter}, \emph{Threshold}, \emph{Join}, and \emph{Similarity Search} present greater difficulty, as they demand multi-step logical composition, implicit schema linking, or pattern-based retrieval.

\textbf{Biomedical reasoning categories.}  
To probe scientific reasoning, we classified \name{} queries into three reasoning categories reflecting processes typical of biomedical experts:

\begin{enumerate}[itemsep=0pt, parsep=0pt,topsep=0pt]
    \item \textbf{Operationalizing implicit scientific conventions:} Queries often invoke domain-specific concepts (e.g., “significantly associated SNPs”) that imply non-obvious thresholds, such as $p < 5 \times 10^{-8}$ for GWAS hits, or directionality from beta coefficients. These conventions are rarely explicit in schemas and must be inferred.
    
    \item \textbf{Incorporating missing contextual knowledge:} Experts often incorporate auxiliary data (e.g., drug approval status or clinical trial phase) even when not directly mentioned. For instance, determining if a drug is “approved” for a condition requires disambiguating indication-specific trial phases beyond any binary “approved” flag.

    \item \textbf{Executing complex multi-hop reasoning workflows:} Many questions in \name{} require chaining relational operations across multiple tables. For example, “Which tissues are genes associated with Parkinson’s disease most significantly expressed in?” requires a four-step inference over gene–disease, gene–expression, tissue annotations, and statistical ranking. LLMs often struggle to translate such multi-hop logic into valid, executable SQL.
\end{enumerate}

Additional biomedical reasoning categories and their distribution are provided in Appendix~\ref{subsec:bio_reasoning} (Table~\ref{tab:bio_reasoning_categories}, Figure~\ref{fig:bio_reasoning_distribution}).

\section{Experiments}
The goal of our experiments is to assess how well LLMs can translate biomedical natural language questions from \name{} into accurate and executable BigQuery SQL queries. We evaluate models across prompting configurations and interaction paradigms on a representative test set of 546 questions, sampled to preserve the expert-authored template distribution of the full benchmark, using SQL execution accuracy and natural-language answer quality as evaluation metrics.

\subsection{Experimental Setup}
\label{subsec:experiments}
\textbf{Models.} We evaluate a range of state-of-the-art open-source and proprietary LLMs. Open-source models include \textbf{Llama-3.1} (70B, 405B) and \textbf{Qwen-2.5-Coder} (14B, 32B). Proprietary models include the \textbf{GPT} family (GPT-4o, GPT-o3-mini, GPT-5.2), the \textbf{Gemini} family (Gemini-2.0-Flash, Gemini-3-Pro), and the \textbf{Claude} family (Claude-3.7-Sonnet, Claude-4.5-Opus). This selection spans a diverse range of parameter scales, computational cost profiles, and architectural design philosophies.

\textbf{Isolated SQL generation.}
We first assess models in a single-turn text-to-SQL setting. Each model receives a baseline prompt containing: (1) the natural language biomedical question, (2) the database schema describing tables, columns, and relationships, and (3) simple instructions on structuring the SQL query to remain consistent with the database schema.


We then vary prompt structure along several axes: (1) adding sample table rows (\textit{3-rows}, \textit{5-rows}), (2) adding few-shot examples (\textit{1-shot}, \textit{3-shot}, \textit{5-shot}, \textit{10-shot}, \textit{20-shot}, \textit{40-shot}), (3) adding domain-specific instructions (e.g., statistical thresholds via \textit{stat-instruct}), and (4) a combined variant of the \textit{3-rows}, \textit{3-shot}, and \textit{stat-instruct} prompts (\textit{combo}). Prompt templates are provided in Appendix~\ref{subsec:baseline_prompts}.

\textbf{Interaction paradigms.} Beyond single-turn prompts, we investigate multi-step paradigms that allow iterative reasoning and query refinement. These systems can request schema details, propose intermediate queries, and update their approach before presenting a final SQL query. These paradigms were evaluated using GPT and Gemini models. We experiment with four paradigms: (1) \textbf{ReAct:} A prompt-orchestrated approach where schema validation, syntax checking, and other external tools are invoked within a multi-step SQL generation process~\citep{yao2023react}. The ReAct prompts used are detailed in Appendix \ref{subsec:react_prompts}. (2) \textbf{Schema Indexing:} Schema descriptions are dynamically retrieved using LlamaIndex to support contextual grounding and table selection. (3) \textbf{DAIL-SQL:} We adapt DAIL-SQL \citep{gao2023text} for use on BiomedSQL. DAIL-SQL is a state-of-the-art text-to-SQL solution that retrieves relevant example SQL queries based on the question and injects them into the prompt for more accurate query generation. It is consistently near the top of leaderboards for popular benchmarks like \textsc{Spider} and BIRD. (4) \textbf{Multi-step query refinement:} We implement an iterative text-to-SQL architecture, called BMSQL, where an initial query is refined through feedback loops based on intermediate results or execution errors, emulating expert-level query refinement. The implementation details of BMSQL are shown in Appendix \ref{subsec:bmsql_prompts}.

\textbf{SQL execution metrics.} We report three SQL performance metrics:
\textbf{Execution Accuracy (EX)}, \textbf{Jaccard Index (JAC)}, and \textbf{Syntax Error Rate (SER)}. EX is a widely used text-to-SQL metric~\citep{spider2018, Li2023BIRD}. It is computed as a binary metric for each question: a generated query receives a score of 1 only if its executed output matches the gold execution results exactly, and 0 otherwise. Dataset-level EX is the proportion of questions receiving a score of 1. We adapt EX for our use case to measure row-wise set equality, comparing the set of UUIDs returned in the case of \verb|SELECT| queries or the set of numeric values returned in the case of \verb|COUNT| and other calculation queries. JAC~\citep{costa2021further}, or intersection over union, is a metric for gauging the similarity of two sets. It tells us how close the LLM-generated SQL query results are to the ground-truth. Unlike EX, JAC will still credit a query that returns slightly more or less rows than the ground-truth, making it a more lenient metric. Finally, SER is simply the proportion of questions in the evaluation set for which the LLM-generated SQL query was not executable.

\textbf{BioScore.} To evaluate natural language responses, we adopt BioScore~\citep{bianchi2026cardbiomedbench}, an LLM-as-a-judge metric computed using GPT-4o. BioScore includes: (1) \textbf{Response Quality Rate (RQR):} The proportion of factually correct answers relative to the gold-standard answer. (2) \textbf{Safety Rate (SR):} The proportion of abstentions among all incorrect or abstained answers, which measures a model’s ability to abstain when uncertain.
    
All metric definitions and prompts are provided in Appendix~\ref{subsec:metrics}. A correlation analysis between the SQL execution and BioScore metrics is presented in Appendix~\ref{subsec:corr}. To mitigate concern over the use of LLM-as-a-judge metrics, a correlation analysis between LLM-generated and domain expert-generated BioScores is presented in Appendix~\ref{subsec:corr_bioscore}. 100 LLM-generated responses were sampled from a variety of different tested models and interaction paradigms. Each of these 100 responses was graded by both a domain expert and GPT-4o. The Spearman correlation between these two sets of per-response scores is 0.89 ($p = 7.40e^{-36}$), and Table \ref{tab:bioscore_comparison} summarizes the resulting score distributions. This high correlation gives us confidence that the LLM-as-a-judge metrics used in our evaluations are stable and accurate.

\textbf{Domain Expert Baseline.} 
Two biomedical analysts independently completed a randomly sampled subset of 20 questions drawn from the 546-question test set. For each question, the analysts were provided with the same database schema and contextual information available to the models and generated a SQL query, execution output, and natural-language answer. We calculated EX, JAC, and RQR separately for each analyst and report the mean across experts. Because this baseline is estimated from 20 questions, it carries substantial uncertainty (95\% CI on EX $\sim$ [68\%, 99\%]) and should be read as an approximate reference. The model–expert gap is nonetheless directional, holding even at the lower bound of this interval. The analysts' responses were not adjudicated or combined into consensus answers, because the objective was to estimate independent expert task performance rather than create additional benchmark annotations; accordingly, inter-annotator agreement was not computed. SR is not applicable because the quiz did not permit abstention, and SER was zero because both analysts produced executable BigQuery SQL for all questions.


\subsection{Evaluation Results}
\label{subsec:eval}
\textbf{LLMs struggle with scientific reasoning in SQL generation.} 
Table~\ref{tab:model_performance_fixed} shows that even top-performing models such as Gemini-3-Pro (EX = 58.1\%, JAC = 62.4\%, RQR = 81.8\%) fall short of domain expert performance (90-95\%). Among open-weight models, despite their small size, Qwen-2.5-Coder-32B achieves competitive EX (40.8\%) and Qwen-2.5-Coder-14B attains strong RQR (62.1\%), outperforming much larger Llama models. Qwen-2.5-Coder-32B has the highest SR of any baseline model (61.0\%), though its elevated SER (15.7\%) means some of this reflects abstention due to failed queries rather than calibrated uncertainty.

\begin{table}[htbp]
    \small
    \centering
    \caption{State-of-the-art LLMs struggle with scientific reasoning-based text-to-SQL tasks (*Domain expert baselines not applicable for SR and token count as described in \S\ref{subsec:experiments}).}
    \label{tab:model_performance_fixed}
    \sisetup{table-parse-only}
    \renewcommand{\arraystretch}{1.2}
    \setlength{\tabcolsep}{3.5pt}
    \begin{tabularx}{\linewidth}{%
        >{\raggedright\arraybackslash}p{3.0cm}
        >{\centering\arraybackslash}p{1.6cm}
        >{\centering\arraybackslash}p{1.6cm}
        >{\centering\arraybackslash}p{1.7cm}
        >{\centering\arraybackslash}p{1.5cm}
        >{\centering\arraybackslash}p{1.5cm}
        >{\centering\arraybackslash}p{1.3cm}}
        \toprule
        \makecell{\textbf{Model}} &
        \makecell{\textbf{EX (\%)}}$\uparrow$ &
        \makecell{\textbf{JAC (\%)}}$\uparrow$ &
        \makecell{\textbf{RQR (\%)}}$\uparrow$ &
        \makecell{\textbf{SR (\%)}}$\uparrow$ &
        \makecell{\textbf{SER (\%)}}$\downarrow$ &
        \makecell{\textbf{\# Tokens}}
        \\
        \midrule
        \rowcolor{highlightrow}\extracolsep{0pt}
        Domain Expert & 90.0 & 90.0 & 95.0 & NA* & 0 & NA* \\ 
        GPT-4o               & 46.9 \pmval{4.2} & 54.7 \pmval{3.8} & 71.2 \pmval{3.8} & 26.1 \pmval{3.7} & 1.3 \pmval{0.9} & 3,689 \\
        GPT-o3-mini          & 53.5 \pmval{4.2} & 60.4 \pmval{3.8} & 73.3 \pmval{3.7} & 29.4 \pmval{3.8} & 0.2 \pmval{0.4} & 3,942\\
        GPT-5.2          & 48.5 \pmval{4.2} & 54.6 \pmval{3.9} & 75.3 \pmval{3.6} & 35.6 \pmval{4.0} & 2.7 \pmval{2.7} & 4,598 \\
        Gemini-2.0-Flash     & 33.7 \pmval{4.0} & 37.0 \pmval{3.9} & 71.1 \pmval{3.8} & 27.2 \pmval{3.7} & 4.2 \pmval{1.7} & 3,692 \\
        Gemini-3-Pro & \textbf{58.1 \pmval{4.3}} & \textbf{62.4 \pmval{3.7}} & \textbf{81.8 \pmval{3.0}} & 52.4 \pmval{4.1} & \textbf{0.0 \pmval{0.0}} & 3,136 \\
        Claude-3.7-Sonnet & 45.4 \pmval{4.2} & 49.8 \pmval{4.0} & 69.8 \pmval{3.8} & 43.0 \pmval{4.1} & 1.6 \pmval{1.1} & 3,805\\
        Claude-4.5-Opus & 54.8 \pmval{4.2} & 59.7 \pmval{3.9} & 80.6 \pmval{3.3} & 35.8 \pmval{4.0} & \textbf{0.0 \pmval{0.0}} & 4,197\\
        Qwen-2.5-Coder-14B   & 37.0 \pmval{4.0} & 32.4 \pmval{3.9} & 62.1 \pmval{4.1} & 42.5 \pmval{4.1} & 11.0 \pmval{2.6} & 3,453\\
        Qwen-2.5-Coder-32B   & 40.8 \pmval{4.1} & 44.4 \pmval{4.0} & 58.2 \pmval{4.1} & \textbf{61.0 \pmval{4.1}} & 15.7 \pmval{3.1} & 3,612 \\
        Llama-3.1-70B        & 34.4 \pmval{4.0} & 39.8 \pmval{3.9} & 57.0 \pmval{4.1} & 37.0 \pmval{4.0} & 6.0 \pmval{2.0} & 3,547 \\
        Llama-3.1-405B       & 38.1 \pmval{4.1} & 42.5 \pmval{4.0} & 57.9 \pmval{4.1} & 41.7 \pmval{4.1} & 4.6 \pmval{1.7} & 3,456\\
        \bottomrule
    \end{tabularx}
\end{table}

\textbf{Few-shot examples provide performance gains.}
Table~\ref{tab:model_performance_varied} in Appendix~\ref{subsec:prompt_experiments} shows that the \textit{10-shot} prompt yields the best improvement for GPT-o3-mini ($\Delta$EX=+7.8\%, $\Delta$JAC=+7.3\%, $\Delta$RQR=+7.0\%). Passing the model more example queries beyond 10, including the full set of 40 template queries, had minimal effect on performance. Passing table rows alone showed little benefit, suggesting schema understanding matters more than content memorization.

\textbf{Interaction paradigms yield mixed results.} 
Table~\ref{tab:model_performance_interaction} shows that schema indexing underperforms in both EX and RQR, likely due to its use of simple table descriptions and lightweight grounding. It exhibits the best SR (e.g., Index-GPT-4o = 66.9\%), but this co-occurs with the highest SER (27.5\%), so part of its apparent safety stems from abstaining due to query errors. ReAct marginally improves EX for GPT variants but does not perform consistently across models. This suggests that ReAct-style prompts may need model-specific tuning to optimize performance. DAIL-SQL shows strong performance, rivaling that of BMSQL on all three models tested. However, it is important to note that even this state-of-the-art text-to-SQL approach still trails expert-level EX performance by $\sim$30 percentage points.

\textbf{BMSQL is the strongest configuration overall.} 
BMSQL-GPT-o3-mini achieves the highest EX (62.6\%) and JAC (69.2\%) of any system we tested, though its margin over DAIL-SQL-GPT-o3-mini (61.2\% EX) falls within confidence intervals. BMSQL does not dominate uniformly: paired with Gemini it reaches 55.9\% EX, below both DAIL-SQL-GPT-o3-mini and the Gemini-3-Pro baseline. BMSQL and DAIL-SQL are best read as comparable top performers rather than one clearly beating the other. On answer quality, BMSQL-Gemini reaches 84.6\% RQR, its closest approach to the expert baseline, though it still trails the 95.0\% mark by roughly ten points.

\begin{table}[htbp]
    \small
    \centering
    \caption{Complex interaction paradigms provide mixed performance (*Gemini-2.0-Flash is the Gemini model used for these experiments).}
    \label{tab:model_performance_interaction}
    \sisetup{table-parse-only}
    \renewcommand{\arraystretch}{1.2}
    \setlength{\tabcolsep}{3.5pt}
    \begin{tabularx}{\linewidth}{%
        >{\raggedright\arraybackslash}p{3.5cm}
        >{\centering\arraybackslash}p{1.5cm}
        >{\centering\arraybackslash}p{1.5cm}
        >{\centering\arraybackslash}p{1.5cm}
        >{\centering\arraybackslash}p{1.5cm}
        >{\centering\arraybackslash}p{1.5cm}
        >{\centering\arraybackslash}p{1.3cm}}
        \toprule
        \makecell{\textbf{Model}} &
        \makecell{\textbf{EX (\%)}}$\uparrow$ &
        \makecell{\textbf{JAC (\%)}}$\uparrow$ &
        \makecell{\textbf{RQR (\%)}}$\uparrow$ &
        \makecell{\textbf{SR (\%)}}$\uparrow$ &
        \makecell{\textbf{SER (\%)}}$\downarrow$ &
        \makecell{\textbf{\# Tokens}}
        \\
        \midrule
        ReAct-GPT-4o & 49.6 \pmval{4.2} & 57.9 \pmval{3.8} & 67.2 \pmval{3.9} & 8.9 \pmval{2.4} & \textbf{0.0 \pmval{0.0}} & 14,286 \\
        ReAct-GPT-o3-mini & 56.2 \pmval{4.2} & 64.8 \pmval{3.6} & 73.6 \pmval{3.7} & 13.2 \pmval{2.8} & \textbf{0.0 \pmval{0.0}} & 13,317 \\
        ReAct-Gemini* & 48.9 \pmval{4.2} & 56.6 \pmval{3.8} & 60.4 \pmval{4.1} & 10.2 \pmval{2.5} & \textbf{0.0 \pmval{0.0}} & 13,205 \\
        Index-GPT-4o & 25.5 \pmval{3.6} & 28.3 \pmval{3.6} & 44.1 \pmval{4.2} & \textbf{66.9 \pmval{3.9}} & 27.5 \pmval{3.7} & 1,110 \\
        Index-GPT-o3-mini & 27.1 \pmval{3.7} & 30.6 \pmval{3.7} & 44.1 \pmval{4.1} & 47.5 \pmval{4.2} &  2.0 \pmval{0.1} & 1,899 \\
        Index-Gemini* & 46.1 \pmval{4.2} & 48.5 \pmval{4.1} & 54.2 \pmval{4.2} & 59.6 \pmval{4.1} & 8.1 \pmval{2.3} & 787\\
        DAIL-SQL-GPT-4o & 54.8 \pmval{4.2} & 58.1 \pmval{3.4} & 75.5 \pmval{3.4} & 63.4 \pmval{4.0} & 6.6 \pmval{2.1} & 3,624 \\
        DAIL-SQL-GPT-o3-mini & 61.2 \pmval{4.1} & 63.6 \pmval{4.0} & 81.4 \pmval{3.3} & 42.1 \pmval{4.1} &  \textbf{0.0 \pmval{0.0}} & 3,318 \\
        DAIL-SQL-Gemini* & 53.1 \pmval{4.2} & 58.8 \pmval{3.4} & 82.8 \pmval{3.1} & 30.6 \pmval{3.7} & \textbf{0.0 \pmval{0.0}} & 3,185\\
        BMSQL-GPT-4o & 60.4 \pmval{4.1} & 67.2 \pmval{3.6} & 79.8 \pmval{3.4} & 64.5 \pmval{4.0} & 4.9 \pmval{1.8} & 32,819 \\
        BMSQL-GPT-o3-mini & \textbf{62.6 \pmval{4.1}} & \textbf{69.2 \pmval{3.6}} & 83.2 \pmval{3.1} & 38.0 \pmval{4.1} & 2.6 \pmval{1.2} & 39,470 \\
        BMSQL-Gemini* & 55.9 \pmval{4.2} & 61.3 \pmval{3.9} & \textbf{84.6 \pmval{3.0}} & 32.1 \pmval{3.9} & 0.2 \pmval{0.4} & 22,045 \\
        \bottomrule
    \end{tabularx}
\end{table}

\section{Analysis}
\label{section:analysis}
To better understand model behavior beyond aggregate metrics, we analyze performance across SQL task types and evaluate the effects of increasing the size of the database.

\begin{figure}[htbp]
    \centering
    \includegraphics[width=0.9\textwidth]{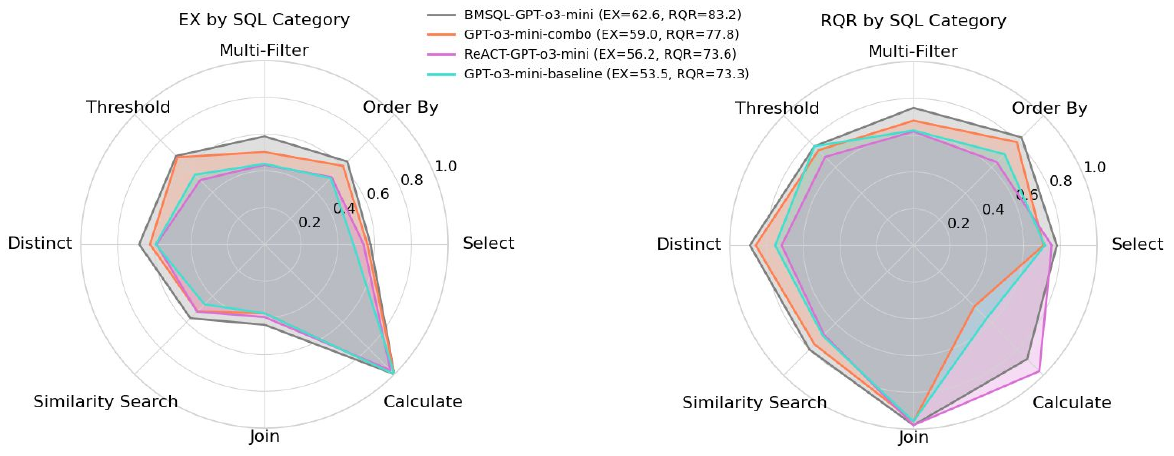}
    \caption{Distribution of performance across SQL categories for GPT-o3-mini in terms of EX (left) and RQR (right), across four prompting strategies and interaction paradigms.}
    \label{fig:performance_distributions}
\end{figure}

\textbf{Performance across SQL categories.}  
Figure~\ref{fig:performance_distributions} presents radar plots showing the distribution of EX and RQR across SQL categories, as defined in \S\ref{sec:dataset_analysis}. We evaluate GPT-o3-mini across four settings: (1) baseline prompt, (2) \textit{combo} prompt, (3) ReAct prompting, and (4) BMSQL. 

For EX, performance across SQL categories remains relatively stable for all prompting strategies. As anticipated, models struggle most with \textit{Join}, \textit{Similarity Search}, and \textit{Multi-Filter} queries, which require multi-table reasoning, implicit filtering, or fuzzy matching. Surprisingly, \textit{Select} queries also show mid-range performance; however, this category includes a large portion of questions in BiomedSQL, so mean-level performance is expected. For RQR, BMSQL exhibits the most balanced performance across categories, likely due to its ability to:
(1) apply domain-specific instructions (e.g., p-value thresholds, trial status),
(2) compare thresholded vs. unthresholded results,
(3) refine queries via execution feedback. This reinforces the benefits of multi-step pipelines in biomedical reasoning.

\textbf{Common SQL errors.} In order to determine the kinds of mistakes the LLMs are making in generating the SQL queries, we defined the following five error categories: (1) incorrect tables, (2) missing threshold, (3) incorrect threshold, (4) incorrect aggregations, and (5) syntax error. More precise definitions of each error category are presented in Appendix~\ref{subsec:sql_errors}, along with the counts of each error type for six of the top-performing models in Table~\ref{tab:sql_errors}. Incorrect table selection is the most common error in the LLM-generated SQL queries, followed by missing or incorrect application of statistical thresholds. The systems also exhibit complementary error profiles: ReAct performs particularly well on table selection and produces no syntax errors, whereas BMSQL’s primary advantage lies in correctly applying domain-specific statistical thresholds. These results highlight the difficulty of distinguishing among closely related tables and the limitations of frontier LLMs in applying biomedical analytical constraints, even when the correct thresholds are provided.


\textbf{Effect of inference-time compute.}  
We next evaluate how model performance changes when given the opportunity to revise outputs over multiple reasoning rounds. Specifically, we allow BMSQL-GPT-o3-mini to examine its initial SQL query, execution result, and answer, then choose to perform up to two additional passes if the outputs appear insufficient. Table~\ref{tab:inference_time_compute} shows that increasing inference steps yields marginal gains. From 1-pass to 3-pass, EX remains flat (62.6\% $\rightarrow$ 61.7\%), while RQR increases slightly (83.2\% $\rightarrow$ 85.5\%). Elimination of syntax errors on the second and third pass (SER = 0.0\%) suggests that most corrections involve fixing syntax rather than changing the query logic. BMSQL rarely invokes a third pass, as seen in the minor token increase between the 2-pass and 3-pass settings.

\begin{table}[htbp]
    \small
    \centering
    \caption{Inference time compute has little effect on performance of BMSQL-GPT-o3-mini.}
    \label{tab:inference_time_compute}
    \sisetup{table-parse-only}
    \renewcommand{\arraystretch}{1.2}
    \setlength{\tabcolsep}{5.4pt}
    \begin{tabularx}{\linewidth}{%
        >{\raggedright\arraybackslash}p{1.1cm}
        >{\centering\arraybackslash}p{1.7cm}
        >{\centering\arraybackslash}p{1.7cm}
        >{\centering\arraybackslash}p{1.8cm}
        >{\centering\arraybackslash}p{1.7cm}
        >{\centering\arraybackslash}p{1.8cm}
        >{\centering\arraybackslash}p{1.6cm}}
        \toprule
        \makecell{\textbf{Model}} &
        \makecell{\textbf{EX (\%)}}$\uparrow$ &
        \makecell{\textbf{JAC (\%)}}$\uparrow$ &
        \makecell{\textbf{RQR (\%)}}$\uparrow$ &
        \makecell{\textbf{SR (\%)}}$\uparrow$ &
        \makecell{\textbf{SER (\%)}}$\downarrow$ &
        \makecell{\textbf{\# Tokens}}
        \\
        \midrule
        1-pass & \textbf{62.6 \pmval{4.1}} & 69.2 \pmval{3.6} & 83.2 \pmval{3.1} & \textbf{38.0 \pmval{4.1}} & 2.6 \pmval{1.2} & 39,470 \\
        2-pass & 62.1 \pmval{4.1} & \textbf{69.4 \pmval{3.5}} & 84.2 \pmval{3.1} & 29.1 \pmval{3.8} & \textbf{0.0 \pmval{0.0}} & 53,773 \\
        3-pass & 61.7 \pmval{4.1} & 69.2 \pmval{3.5} & \textbf{85.5 \pmval{2.9}} & 36.7 \pmval{4.0} & 0.0 \pmval{0.0} & 55,948 \\
        \bottomrule
    \end{tabularx}
\end{table}

\textbf{Effect of database size.} BiomedSQL was created from a database of ten tables. Although this is small relative to other benchmarks, the tables themselves contain up to 72 million rows and 31 columns, requiring models to identify the correct columns and apply question-specific statistical thresholds. To evaluate performance under schema expansion, we added ten new tables containing additional disease-specific GWAS results, drug mechanism-of-action data, and population-level allele frequencies. We evaluated GPT-o3-mini on this 20-table schema using four settings: (1) baseline prompting, (2) \textit{10-shot} prompting, (3) \textit{combo} prompting, and (4) BMSQL (Table~\ref{tab:big_schema}, Appendix~\ref{subsec:big_schema}). On the original schema, 10-shot prompting performs comparably to BMSQL in execution accuracy (61.3\% vs. 62.6\%) while using approximately one-seventh as many tokens, demonstrating that single-call few-shot prompting is a strong and efficient baseline. On the larger schema, 10-shot prompting achieves 53.8\% EX and 58.5\% JAC, compared with 58.4\% EX and 65.1\% JAC for BMSQL. BMSQL therefore exhibits a smaller decline in performance under schema expansion: EX decreases by 4.2 percentage points for BMSQL and 7.5 points for 10-shot prompting, while JAC decreases by 4.1 and 9.2 points, respectively. These differences are within the uncertainty of the estimates and should therefore be interpreted as an observed trend rather than statistically definitive evidence of greater robustness. Within this experimental setting, the results suggest that BMSQL’s iterative schema inspection, query refinement, and domain-specific filtering may become more beneficial as schema complexity increases, although at substantially greater token cost.





\section{Discussion and Limitations}
\label{sec:discussion}
\textbf{Use of template questions.} BiomedSQL was constructed from a set of 40 template questions. While templating can produce more homogeneous samples than real-world settings, it is common in the text-to-SQL community \citep{gao2023text, gan2021towards, saparina2024ambrosia}. 
Popular text-to-SQL benchmarks with large or no template sets often crowdsource their tasks (e.g., BIRD~\citep{Li2023BIRD}), which \textbf{only works when questions come from general domains that crowdworkers can reliably annotate.} Conversely, \textbf{BiomedSQL covers a highly technical domain that cannot be easily crowdsourced}. Instead, all questions and SQL queries were authored by domain experts (co-authors of this work) who are advanced biomedical data scientists and regularly implement complex bioinformatics workflows. Their expertise is essential for generating technically meaningful SQL annotations, resulting in a detail-oriented dataset construction process.

Recent work further demonstrates that templating does not trivialize evaluation. In GSM-Symbolic \citep{mirzadeh2024gsm}, \textbf{regenerating questions through simple templates} (e.g., replacing only surface-level entities such as names or numeric values) \textbf{still causes LLMs to exhibit significant accuracy drops} and increased performance variance relative to the original benchmark, even though the underlying question structure and required reasoning remain unchanged. These results indicate that current models do not reliably generalize their reasoning across equivalent instantiations of the same template. This sensitivity to lexical change may also be intensified in a domain like biomedicine. Entities like \textbf{genes, diseases, and drug names appear at highly heterogeneous frequencies in model pretraining corpora}. The effects of this long-tail knowledge phenomenon have been well-studied, showing that LLM performance disproportionately degrades on rare or infrequent entities \citep{razeghi2022impact, kandpal2023large}. As a result, \textbf{substituting rare biomedical entities within the same template can introduce a meaningful distribution shift} and negatively impact performance, even when the logical form of the query remains constant. We also cite the difficulty of the benchmark task despite our use of templates, as top models still trail expert-level EX performance by $\sim$30 percentage points.


\textbf{Multiple valid SQL solutions.}  
While gold SQL queries in \name{} were authored by domain experts and independently verified by analysts, they do not represent the only correct way to retrieve relevant data for a given question. Since biomedical questions often permit multiple semantically valid formulations, we evaluate models using a combination of execution-based metrics (EX, JAC) and LLM-judged response quality (RQR).



\textbf{Use of BigQuery.} 
While \name{}'s reliance on BigQuery may limit compatibility with other work, we view this as an important design decision. Cloud-native SQL dialects are increasingly common in production systems, especially biomedical research pipelines. Evaluating LLMs on BigQuery brings new challenges, including vendor-specific functions and syntax, that have been underexplored in the research community.


\textbf{General-purpose coding agents.} Because our experiments target controlled text-to-SQL generation, we do not evaluate open-ended coding agents such as Claude Code or Gemini CLI, which iteratively generate, execute, and debug SQL. Such agents could be scored on the same final-output metrics, but would introduce confounds like tool orchestration, execution environments, and debugging budgets. Evaluating them is therefore an important extension, and our results should not be read as evidence that bespoke agents are necessary to match the strongest results on BiomedSQL.

\textbf{Future directions.} We plan to expand our set of template questions to cover the experimental 20-table schema. We also plan to convert the benchmark to SQLite for easier integration with general-purpose text-to-SQL systems such as DIN-SQL~\citep{pourreza2023din} and CHESS~\citep{talaei2024chess}.




\section{Conclusion}
\label{sec:conclusion}

We present \name{}, the first large-scale text-to-SQL benchmark designed to evaluate scientific reasoning during SQL generation in the biomedical domain. Across models and interaction paradigms, \name{} remains challenging, with execution accuracy and answer quality trailing domain experts. Our analyses identify incorrect table selection and the application of domain-specific statistical thresholds as persistent failure modes, while the complementary behavior of ReAct and BMSQL suggests that schema handling and biomedical constraint application remain distinct challenges.

By emphasizing domain conventions, multi-step reasoning, and structured biomedical data, \name{} provides a rigorous testbed for developing more capable and trustworthy systems across broader question families, schemas, SQL dialects, and agentic settings. The benchmark and accompanying resources provide a foundation for systems that can broaden access to biomedical knowledge and accelerate scientific discovery.

\section*{Acknowledgments}
This research was supported in part by the Intramural Research Program of the NIH, National Institute
on Aging (NIA), National Institutes of Health, Department of Health and Human Services; project
number Z01 AG000534, as well as the National Institute of Neurological Disorders and Stroke.

This research was supported [in part] by the Intramural Research Program of the National Institutes
of Health (NIH). The contributions of the NIH author(s) are considered Works of the United States
Government. The findings and conclusions presented in this paper are those of the author(s) and do
not necessarily reflect the views of the NIH or the U.S. Department of Health and Human Services.

Some authors’ participation in this project was part of a competitive contract awarded to DataTecnica
LLC by the National Institutes of Health to support open science research.

This work utilized the computational resources of the NIH HPC Biowulf cluster (\url{https://hpc.nih.gov}).

\section*{Ethics Statement}
BiomedSQL is designed to improve LLMs’ ability to support biomedical discovery, especially for non-technical researchers querying structured data. It is built from public, non-identifiable data. We note that SQL queries generated by LLMs may contain subtle errors that produce plausible but incorrect results. In biomedical settings, where research findings may inform downstream clinical decision-making, we recommend that all model-generated queries and their outputs be reviewed by domain experts before use. Beyond this and the ethical considerations broadly associated with foundation models, we do not foresee negative societal impacts unique to this work.

\section*{Reproducibility Statement}
The complete benchmark, including all questions and the source tables required to reconstruct the database, is publicly available at
\url{https://huggingface.co/datasets/NIH-CARD/BiomedSQL}.

We release self-contained code for reproducing the main experimental results at
\url{https://github.com/NIH-CARD/biomedsql}. The repository includes scripts for constructing the BigQuery database, all prompts used in our experiments, and implementations of each prompting configuration and interaction paradigm. Instructions for environment setup and running the experiments are provided in the repository. Table~\ref{tab:compute_resources} reports the computational resources required for each experiment.

\bibliography{ref_daniel}
\bibliographystyle{colm2026_conference}

\newpage
\appendix
\section{Appendix}


\subsection{Database tables.}
\label{subsec:tables}
This section provides schema details and a short description of the ten core tables in the BiomedSQL BigQuery database. Database Tables 1-2, 6, and 9-10 are generated from sources that are under the CC0 1.0 License or are otherwise designated to the public domain. Database Tables 7-8 are generated from sources that are under either the CC0 1.0 License (Open Targets) or the CC Attribution-ShareAlike 3.0 Unported license (ChEMBL). Database Tables 3 and 5 are generated from sources under the CC BY 4.0 License; Database Table 4 is under CC BY-NC 4.0.

\begingroup
  \setlength{\abovecaptionskip}{12pt}
  \setlength{\belowcaptionskip}{0pt}
  \captionof{database}{Alzheimer's Disease GWAS (21.1M rows, 13 columns).}
  \label{dbtab:adgwas}%
\endgroup
\begin{promptbox}
Table: AlzheimerDisease_CombinedGeneData_UUID
Description: Summary statistics from the largest publicly available GWAS of Alzheimer's Disease in a
European population (*@\citep{bellenguez2022new}@*).
Schema:
  - Name: UUID | Type: STRING | Mode: REQUIRED
  - Name: SNP | Type: STRING | Mode: NULLABLE
  - Name: A1 | Type: STRING | Mode: NULLABLE
  - Name: A2 | Type: STRING | Mode: NULLABLE
  - Name: freq | Type: FLOAT | Mode: NULLABLE
  - Name: b | Type: FLOAT | Mode: NULLABLE
  - Name: se | Type: FLOAT | Mode: NULLABLE
  - Name: p | Type: FLOAT | Mode: NULLABLE
  - Name: chr_37 | Type: INTEGER | Mode: NULLABLE
  - Name: bp_37 | Type: INTEGER | Mode: NULLABLE
  - Name: chr_38 | Type: INTEGER | Mode: NULLABLE
  - Name: bp_38 | Type: INTEGER | Mode: NULLABLE
  - Name: nearestGene | Type: STRING | Mode: NULLABLE
\end{promptbox}

\begingroup
  \setlength{\abovecaptionskip}{12pt}
  \setlength{\belowcaptionskip}{0pt}
  \captionof{database}{Parkinson's Disease GWAS (7.8M rows, 13 columns).}
  \label{dbtab:pdgwas}%
\endgroup
\begin{promptbox}
Table: ParkinsonDisease_CompleteGeneData_No23andMe
Description: Summary statistics from the largest publicly available GWAS of Parkinson's Disease in a
European population (*@\citep{nalls2019parkinsons}@*).
Schema:
  - Name: UUID | Type: STRING | Mode: REQUIRED
  - Name: SNP | Type: STRING | Mode: NULLABLE
  - Name: A1 | Type: STRING | Mode: NULLABLE
  - Name: A2 | Type: STRING | Mode: NULLABLE
  - Name: freq | Type: FLOAT | Mode: NULLABLE
  - Name: b | Type: FLOAT | Mode: NULLABLE
  - Name: se | Type: FLOAT | Mode: NULLABLE
  - Name: p | Type: FLOAT | Mode: NULLABLE
  - Name: chr_37 | Type: INTEGER | Mode: NULLABLE
  - Name: bp_37 | Type: INTEGER | Mode: NULLABLE
  - Name: chr_38 | Type: INTEGER | Mode: NULLABLE
  - Name: bp_38 | Type: INTEGER | Mode: NULLABLE
  - Name: nearestGene | Type: STRING | Mode: NULLABLE
\end{promptbox}

\begingroup
  \setlength{\abovecaptionskip}{12pt}
  \setlength{\belowcaptionskip}{0pt}
  \captionof{database}{Alzheimer's Disease Gene Pathway Associations (542 rows, 5 columns).}
  \label{dbtab:adpath}%
\endgroup
\begin{promptbox}
Table: AlzheimerDisease_GeneAssoc_Pathways_UUID
Description: Summary statistics from a pathway-level analysis of gene sets in Alzheimer's Disease
(*@\citep{zhang2021alzheimers}@*).
Schema:
  - Name: UUID | Type: STRING | Mode: REQUIRED
  - Name: genes | Type: STRING | Mode: NULLABLE
  - Name: size | Type: INTEGER | Mode: NULLABLE
  - Name: statistic | Type: FLOAT | Mode: NULLABLE
  - Name: p | Type: FLOAT | Mode: NULLABLE
\end{promptbox}

\newpage
\begingroup
  \setlength{\abovecaptionskip}{0pt}
  \setlength{\belowcaptionskip}{0pt}
  \captionof{database}{Parkinson's Disease Gene Pathway Associations (1,016 rows, 5 columns).}
  \label{dbtab:pdpath}%
\endgroup
\begin{promptbox}
Table: ParkinsonDisease_GeneAssoc_Pathways_UUID
Description: Summary statistics from a pathway-level analysis of gene sets in Parkinson's Disease
(*@\citep{elango2023parkinsons}@*).
Schema:
  - Name: UUID | Type: STRING | Mode: REQUIRED
  - Name: genes | Type: STRING | Mode: NULLABLE
  - Name: size | Type: INTEGER | Mode: NULLABLE
  - Name: statistic | Type: FLOAT | Mode: NULLABLE
  - Name: p | Type: FLOAT | Mode: NULLABLE
\end{promptbox}

\begingroup
  \setlength{\abovecaptionskip}{12pt}
  \setlength{\belowcaptionskip}{0pt}
  \captionof{database}{Neurodegenerative Disease SMR Associations (1.7M rows, 31 columns).}
  \label{dbtab:smr}%
\endgroup
\begin{promptbox}
Table: NeurodegenerativeDiseases_SMR_Genes_Full
Description: SMR results providing functional inferences between genetic variants and six neurodegenerative 
diseases (*@\citep{alvarado2024omicsynth}@*).
Schema:
  - Name: UUID | Type: STRING | Mode: REQUIRED
  - Name: Omic | Type: STRING | Mode: NULLABLE
  - Name: Disease | Type: STRING | Mode: NULLABLE
  - Name: probeID | Type: STRING | Mode: NULLABLE
  - Name: ProbeChr | Type: INTEGER | Mode: NULLABLE
  - Name: Gene | Type: STRING | Mode: NULLABLE
  - Name: Probe_bp | Type: INTEGER | Mode: NULLABLE
  - Name: topSNP | Type: STRING | Mode: NULLABLE
  - Name: topSNP_chr | Type: INTEGER | Mode: NULLABLE
  - Name: topSNP_bp | Type: INTEGER | Mode: NULLABLE
  - Name: A1 | Type: STRING | Mode: NULLABLE
  - Name: A2 | Type: STRING | Mode: NULLABLE
  - Name: Freq | Type: FLOAT | Mode: NULLABLE
  - Name: b_GWAS | Type: FLOAT | Mode: NULLABLE
  - Name: se_GWAS | Type: FLOAT | Mode: NULLABLE
  - Name: p_GWAS | Type: FLOAT | Mode: NULLABLE
  - Name: b_eQTL | Type: FLOAT | Mode: NULLABLE
  - Name: se_eQTL | Type: FLOAT | Mode: NULLABLE
  - Name: p_eQTL | Type: FLOAT | Mode: NULLABLE
  - Name: b_SMR | Type: FLOAT | Mode: NULLABLE
  - Name: se_SMR | Type: FLOAT | Mode: NULLABLE
  - Name: p_SMR | Type: FLOAT | Mode: NULLABLE
  - Name: p_SMR_multi | Type: FLOAT | Mode: NULLABLE
  - Name: p_HEIDI | Type: FLOAT | Mode: NULLABLE
  - Name: nsnp_HEIDI | Type: FLOAT | Mode: NULLABLE
  - Name: topRSID | Type: STRING | Mode: NULLABLE
  - Name: Omic_type | Type: STRING | Mode: NULLABLE
  - Name: Omic_tissue | Type: STRING | Mode: NULLABLE
  - Name: Disease_name | Type: STRING | Mode: NULLABLE
  - Name: Source | Type: STRING | Mode: NULLABLE
  - Name: func_sig | Type: STRING | Mode: NULLABLE
\end{promptbox}

\begingroup
  \setlength{\abovecaptionskip}{12pt}
  \setlength{\belowcaptionskip}{0pt}
  \captionof{database}{Neurodegenerative Disease Allele Frequencies (72.2M rows, 6 columns).}
  \label{dbtab:freq}%
\endgroup
\begin{promptbox}
Table: NeurodegenerativeDisease_AlleleFrequencies_UUID
Description: Allele frequencies from a cohort not containing Alzheimer's or Parkinson's disease cases
(*@\citep{bergstrom2020allele}@*).
Schema:
  - Name: UUID | Type: STRING | Mode: REQUIRED
  - Name: chr_38 | Type: INTEGER | Mode: NULLABLE
  - Name: SNP | Type: STRING | Mode: NULLABLE
  - Name: A1 | Type: STRING | Mode: NULLABLE
  - Name: A2 | Type: STRING | Mode: NULLABLE
  - Name: freq | Type: FLOAT | Mode: NULLABLE
\end{promptbox}

\newpage
\begingroup
  \setlength{\abovecaptionskip}{0pt}
  \setlength{\belowcaptionskip}{0pt}
  \captionof{database}{Drug Gene Targets (6,391 rows, 20 columns).}
  \label{dbtab:dgt}%
\endgroup
\begin{promptbox}
Table: DrugGeneTargets_ComprehensiveAnnotations_updated
Description: Details drug-gene relationships and offers a comprehensive view of drug development pipelines
(*@\citep{opentargets, CHEMBL}@*).
Schema:
  - Name: UUID | Type: STRING | Mode: REQUIRED
  - Name: chemblIdentifier | Type: STRING | Mode: NULLABLE
  - Name: blackBoxWarning | Type: BOOLEAN | Mode: NULLABLE
  - Name: drugName | Type: STRING | Mode: NULLABLE
  - Name: drugMolecularType | Type: STRING | Mode: NULLABLE
  - Name: yearOfFirstApproval | Type: INTEGER | Mode: NULLABLE
  - Name: maxClinicalTrialPhase | Type: INTEGER | Mode: NULLABLE
  - Name: drugHasBeenWithdrawn | Type: BOOLEAN | Mode: NULLABLE
  - Name: drugIsApproved | Type: BOOLEAN | Mode: NULLABLE
  - Name: tradeNames_string | Type: STRING | Mode: NULLABLE
  - Name: drugSynonyms_string | Type: STRING | Mode: NULLABLE
  - Name: linkedDiseasesDrug_string | Type: STRING | Mode: NULLABLE
  - Name: linkedDiseasesCount | Type: INTEGER | Mode: NULLABLE
  - Name: newLinkedTargets_string | Type: STRING | Mode: NULLABLE
  - Name: numberLinkedTargets | Type: INTEGER | Mode: NULLABLE
  - Name: drugDescription | Type: STRING | Mode: NULLABLE
  - Name: drugActionType | Type: STRING | Mode: NULLABLE
  - Name: drugMechanismOfAction | Type: STRING | Mode: NULLABLE
  - Name: tradename_count | Type: INTEGER | Mode: NULLABLE
  - Name: synonyms_count | Type: INTEGER | Mode: NULLABLE
\end{promptbox}

\begingroup
  \setlength{\abovecaptionskip}{12pt}
  \setlength{\belowcaptionskip}{0pt}
  \captionof{database}{Drug Target Indications (1.2M rows, 23 columns).}
  \label{dbtab:dti}%
\endgroup
\begin{promptbox}
Table: DrugTargets_IndicationsAndTherapeuticUses
Description: Links drugs to specific indications, facilitating disease- and target-specific 
therapeutic explorations (*@\citep{opentargets, CHEMBL}@*).
Schema:
  - Name: UUID | Type: STRING | Mode: REQUIRED
  - Name: chemblId | Type: STRING | Mode: NULLABLE
  - Name: drugName | Type: STRING | Mode: NULLABLE
  - Name: tradeName | Type: STRING | Mode: NULLABLE
  - Name: drugType | Type: STRING | Mode: NULLABLE
  - Name: actionType | Type: STRING | Mode: NULLABLE
  - Name: targetType | Type: STRING | Mode: NULLABLE
  - Name: target | Type: STRING | Mode: NULLABLE
  - Name: approvedSymbol | Type: STRING | Mode: NULLABLE
  - Name: approvedName | Type: STRING | Mode: NULLABLE
  - Name: yearOfFirstApproval | Type: INTEGER | Mode: NULLABLE
  - Name: usan_year | Type: FLOAT | Mode: NULLABLE
  - Name: patent_no | Type: STRING | Mode: NULLABLE
  - Name: max_phase_for_ind | Type: FLOAT | Mode: NULLABLE
  - Name: mesh_id | Type: STRING | Mode: NULLABLE
  - Name: mesh_heading | Type: STRING | Mode: NULLABLE
  - Name: efo_id | Type: STRING | Mode: NULLABLE
  - Name: efo_term | Type: STRING | Mode: NULLABLE
  - Name: tradeNames_list | Type: STRING | Mode: NULLABLE
  - Name: tradename_count | Type: INTEGER | Mode: NULLABLE
  - Name: syns_list | Type: STRING | Mode: NULLABLE
  - Name: synonyms_count | Type: INTEGER | Mode: NULLABLE
  - Name: ct | Type: STRING | Mode: NULLABLE
\end{promptbox}

\begingroup
  \setlength{\abovecaptionskip}{12pt}
  \setlength{\belowcaptionskip}{0pt}
  \captionof{database}{Drug Licensing (2,097 rows, 16 columns).}
  \label{dbtab:license}%
\endgroup
\begin{promptbox}
Table: DrugTargets_LicensingAndUses
Description: Licensing, pharmaceutical company, and dosage information for specific drugs (*@\citep{purplebook}@*).
Schema:
  - Name: UUID | Type: STRING | Mode: REQUIRED
  - Name: applicant | Type: STRING | Mode: NULLABLE
  - Name: blaNumber | Type: INTEGER | Mode: NULLABLE
  - Name: tradeName | Type: STRING | Mode: NULLABLE
  - Name: drugName | Type: STRING | Mode: NULLABLE
  - Name: blaType | Type: STRING | Mode: NULLABLE
  - Name: strength | Type: STRING | Mode: NULLABLE
  - Name: dosageForm | Type: STRING | Mode: NULLABLE
  - Name: routeOfAdministration | Type: STRING | Mode: NULLABLE
  - Name: productPresentation | Type: STRING | Mode: NULLABLE
  - Name: marketingStatus | Type: STRING | Mode: NULLABLE
  - Name: licensure | Type: STRING | Mode: NULLABLE
  - Name: submissionType | Type: STRING | Mode: NULLABLE
  - Name: licenseNumber | Type: INTEGER | Mode: NULLABLE
  - Name: productNumber | Type: INTEGER | Mode: NULLABLE
  - Name: center | Type: STRING | Mode: NULLABLE
\end{promptbox}

\newpage
\begingroup
  \setlength{\abovecaptionskip}{0pt}
  \setlength{\belowcaptionskip}{0pt}
  \captionof{database}{Drug Dosages (211k rows, 11 columns).}
  \label{dbtab:dosage}%
\endgroup
\begin{promptbox}
Table: DrugTargets_UsesAndDosages
Description: Dosage, route of administration, and strength information for specific drugs (*@\citep{drugcode}@*).
Schema:
  - Name: UUID | Type: STRING | Mode: REQUIRED
  - Name: productType | Type: STRING | Mode: NULLABLE
  - Name: tradeName | Type: STRING | Mode: NULLABLE
  - Name: drugName | Type: STRING | Mode: NULLABLE
  - Name: dosageForm | Type: STRING | Mode: NULLABLE
  - Name: dosageRoute | Type: STRING | Mode: NULLABLE
  - Name: labelerName | Type: STRING | Mode: NULLABLE
  - Name: activeDosage_strength | Type: STRING | Mode: NULLABLE
  - Name: activeIngredient_strength | Type: STRING | Mode: NULLABLE
  - Name: mechanismOfAction_pharma | Type: STRING | Mode: NULLABLE
  - Name: packageDescription | Type: STRING | Mode: NULLABLE
\end{promptbox}

\subsection{Biomedical reasoning categories.}
\label{subsec:bio_reasoning}
\begin{figure}[htbp]
    \centering
    \includegraphics[width=0.45\textwidth]{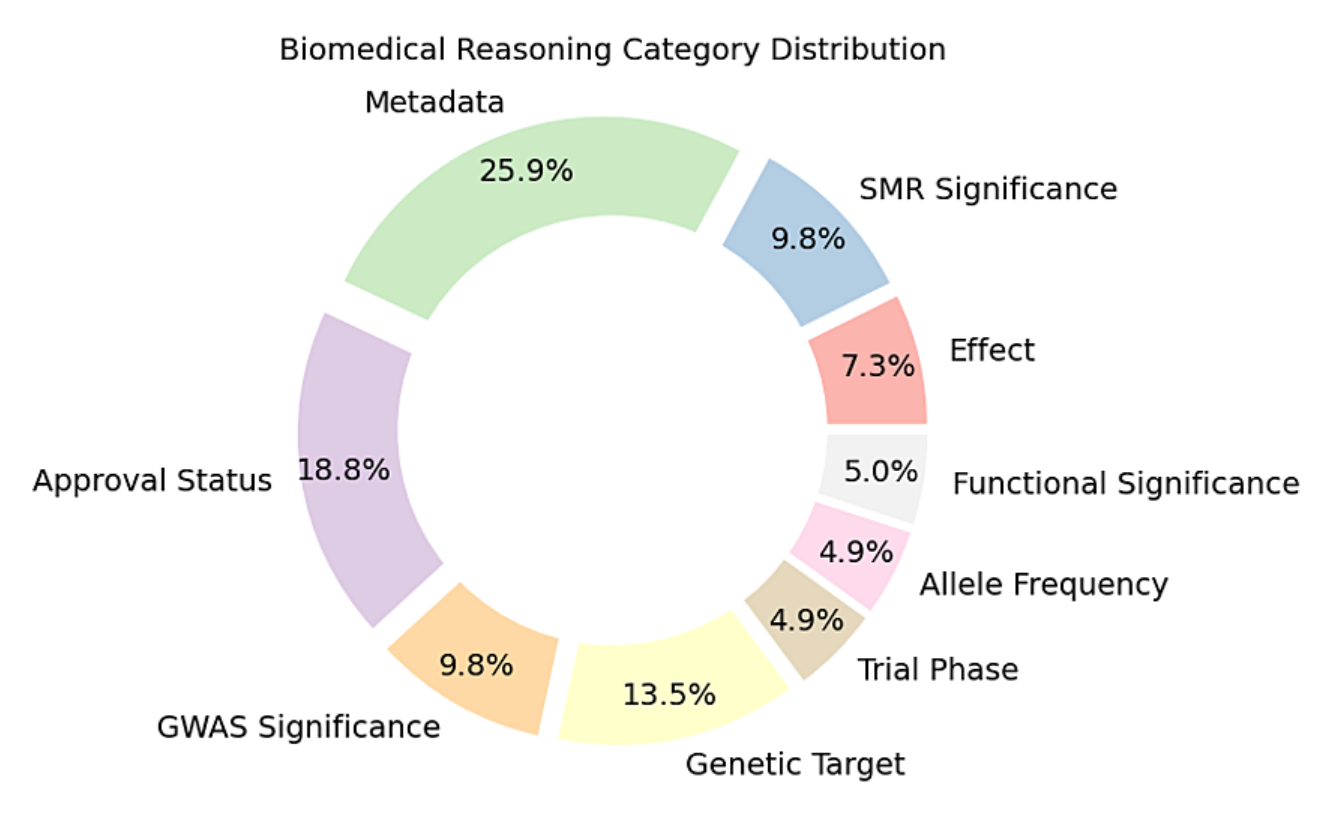}
    \caption{Biomedical reasoning categories.}
    \label{fig:bio_reasoning_distribution}
\end{figure}

Table \ref{tab:bio_reasoning_categories} defines the biomedical reasoning categories that BiomedSQL challenges and Figure \ref{fig:bio_reasoning_distribution} shows their distribution among the set of 68,000 queries. The \textit{GWAS Significance}, \textit{SMR Significance}, and \textit{Functional Significance} categories test the ability of LLMs to operationalize domain-specific statistical significance thresholds. Categories such as \textit{Approval Status}, \textit{Genetic Target}, and \textit{Trial Phase} task LLMs with understanding and applying information about clinical trial phases for specific drugs and indications to generate a correct SQL query.

\begin{table}[htbp]
    \centering
    \small
    \caption{Description of biomedical reasoning categories in \name{}.}
    \label{tab:bio_reasoning_categories}
    \setlength{\tabcolsep}{5pt}
    \begin{tabularx}{\linewidth}{lX}
        \toprule
        \textbf{Biomedical Category} & \textbf{Description} \\
        \midrule
        Approval Status & Retrieves information on the FDA approval status of a drug or indication. \\
        Trial Phase & Retrieves information on the clinical trial phase a drug has reached. \\
        GWAS Significance & Identifies variants that are GWAS significant for a disease ($p < 5e^{-8}$). \\
        SMR Significance & Identifies variants that are SMR significant for a disease ($p < 2.95e^{-6}$). \\
        Functional Significance & Identifies variants that are significant in a particular tissue for a disease. \\
        Effect &  Retrieves the effect size and direction for specific variants. \\
        Genetic Target & Retrieves information on the genetic target of a drug. \\
        Allele Frequency & Calculates allele frequencies for a variant or set of variants. \\
        Metadata & Retrieves general information on a drug or genetic variant. \\
        \bottomrule
    \end{tabularx}
\end{table}

\newpage
\subsection{Isolated SQL generation prompt templates.}
\label{subsec:baseline_prompts}
This section provides details about the prompt engineering approaches used for the isolated SQL query generation experiments.

Prompt \ref{prm:baseline} contains the baseline prompt template that was used. For these experiments, the \textit{db\_schema} variable is replaced with the schema that is detailed in Appendix \ref{subsec:tables}. In the case of the \textit{3-rows}, \textit{5-rows}, and \textit{combo} experiments, \textit{db\_schema} is replaced with a schema that includes the corresponding number of example rows for each table in the database.

Prompt \ref{prm:examples} shows the example queries that were appended to the baseline prompt for use in the \textit{1-shot}, \textit{3-shot}, \textit{5-shot}, and \textit{combo} experiments. For brevity, we have not included the prompt that contains additional example queries for the \textit{10-shot}, \textit{20-shot}, and \textit{40-shot} experiments. Note that the example queries stay the same regardless of the question from BiomedSQL being passed to the model.

Prompt \ref{prm:stats} details the statistical thresholding instructions that were appended to the baseline prompt for use in the \textit{stat-instruct} and \textit{combo} experiments.

Finally, Prompt \ref{prm:response} contains the prompt template that was passed to the LLMs for the generation of a final natural language response based on the question, generated SQL query, and execution results. This prompt was used throughout the isolated SQL generation experiments.

\begin{promptbox}
You are a data analyst and SQL developer experienced with biomedical data in Google BigQuery.
Your task is to translate the user's natural language question into a syntactically correct
Google BigQuery SQL query.

User's Natural Language Question:
{question}

Database Schema:
{db_schema}

Use these guidelines when generating the query:
    1. Review the database schema.
    2. Review the user's question.
    3. Generate a valid Google BigQuery SQL query that answers the question based on the schema.
    4. Always enclose table references in backticks, e.g., `project.dataset.table`.
    5. Make use of BigQuery-specific functions and syntax where appropriate
    (e.g., DISTINCT, aliases, ORDER BY).
    6. Always include the UUID column in your SELECT statements, except in cases of questions where the
    COUNT and GROUP BY functions are needed.
    7. Unless the user explicitly requests a different LIMIT, default your queries to LIMIT 100.
    8. Output ONLY the raw SQL query (no additional commentary or explanations).
    9. Avoid SELECT *; select only the necessary columns to answer the user's query.
    10. Ensure that any disease names that contain an apostrophe in the query are surrounded by
    double quotes (e.g., "Alzheimer's Disease").

Please only return the SQL query in the following format:
```
{{sql_query}}
```
\end{promptbox}
\begingroup
  \setlength{\abovecaptionskip}{6pt}
  \setlength{\belowcaptionskip}{10pt}
  \captionof{prompt}{Baseline prompt template for the isolated SQL generation experiments.}
  \label{prm:baseline}
\endgroup

\newpage
\begin{promptbox}
Below are example BigQuery queries to guide you (for reference only, do not repeat verbatim unless needed
by the user's request):
Example 1:
    SELECT DISTINCT drugName, drugIsApproved, newLinkedTargets_string 
    FROM `bio_sql_benchmark.DrugGeneTargets_ComprehensiveAnnotations_updated`
    WHERE newLinkedTargets_string LIKE "%TUBB %" 
    AND drugIsApproved = TRUE
    LIMIT 100
Example 2:
    SELECT SNP, A1 AS effect_allele, freq AS effect_allele_frequency,
    A2 AS non_effect_allele, 1 - freq AS non_effect_allele_frequency
    FROM `bio_sql_benchmark.AlzheimerDisease_CombinedGeneData_UUID`
    WHERE SNP = 'rs61769339'
    LIMIT 100
Example 3:
    SELECT drugName, newLinkedTargets_string, drugIsApproved
    FROM `bio_sql_benchmark.DrugGeneTargets_ComprehensiveAnnotations_updated`
    WHERE newLinkedTargets_string LIKE '%ACACA%'
    AND drugIsApproved = TRUE
    LIMIT 100
Example 4:
    SELECT topRSID, Disease, Gene, p_SMR_multi, p_HEIDI, b_SMR
    FROM `bio_sql_benchmark.NeurodegenerativeDiseases_SMR_Genes_Full`
    WHERE Disease = 'FTD' AND Gene = 'ORC3' AND p_SMR_multi < 2.95e-6
    LIMIT 100
Example 5:
    SELECT DISTINCT drugName, tradeNames_list, drugType, actionType, target, approvedSymbol, approvedName,
    yearOfFirstApproval, max_phase_for_ind, mesh_heading, efo_term
    FROM `bio_sql_benchmark.DrugTargets_IndicationsAndTherapeuticUses`
    WHERE (LOWER(efo_term) = "acute hepatic porphyria"
    OR LOWER(mesh_heading) = "acute hepatic porphyria") AND LOWER(drugType) = "oligonucleotide"
    AND yearOfFirstApproval > 0 AND max_phase_for_ind = 4.0
    LIMIT 100
\end{promptbox}
\begingroup
  \setlength{\abovecaptionskip}{6pt}
  \setlength{\belowcaptionskip}{10pt}
  \captionof{prompt}{Example queries appended to the baseline prompt for the \textit{n-shot} and \textit{combo} experiments.}
  \label{prm:examples}
\endgroup

\begin{promptbox}
Use the following p-value thresholds for questions about statistical significance:
    1. p < 5e-8 for genome-wide significance.
    2. p_SMR < 2.95e-6 for SMR significance.
    3. p_SMR < 2.95e-6, p_HEIDI > 0.01 for functional significance.
\end{promptbox}
\begingroup
  \setlength{\abovecaptionskip}{6pt}
  \setlength{\belowcaptionskip}{10pt}
  \captionof{prompt}{Statistical thresholding instructions appended to the baseline prompt for the \textit{stat-instruct} and \textit{combo} experiments.}
  \label{prm:stats}
\endgroup

\begin{promptbox}
You are a data analyst and SQL developer experienced with biomedical data in Google BigQuery.
Given the following question, SQL query, and SQL query execution results, please provide a concise answer.
Please do not use any information outside of the SQL query and SQL query execution results in your answer.
Question: 
{question}
SQL Query:
{sql_query}
Execution Results:
{execution_results}
\end{promptbox}
\begingroup
  \setlength{\abovecaptionskip}{6pt}
  \setlength{\belowcaptionskip}{10pt}
  \captionof{prompt}{Natural language response prompt template for the isolated SQL generation experiments.}
  \label{prm:response}
\endgroup

\subsection{ReAct prompt template.}
\label{subsec:react_prompts}
Prompt \ref{prm:react} shows the ReAct-style prompt template used in the interaction paradigm experiments. Similar to the baseline prompt, \textit{db\_schema} is replaced with the schema that is detailed in Appendix \ref{subsec:tables}. \textit{history\_str} is replaced with the reasoning trace from previous steps that the LLM chose to take. We allow the LLM to perform up to 5 iterations within the ReAct loop before a final answer is generated.

\begin{promptbox}
You are an expert SQL agent that uses step-by-step reasoning to answer questions about data in a 
BigQuery database.
        
IMPORTANT: The dataset name is "{dataset_name}". Always qualify table names with this dataset name.
Example: SELECT * FROM {dataset_name}.table_name
        
Follow these steps:
    1. Think about how to translate the question into a SQL query.
    2. Decide which tables and columns are needed.
    3. Write a SQL query with explanatory comments.
    4. Verify the query syntax before executing.
    5. If the query has errors, fix them and try again.
    6. Once the query is successful, explain the results clearly.
    7. Always include the UUID column in your SELECT statements, except in cases of questions where 
    the COUNT and GROUP BY functions are needed.
    8. Unless the user explicitly requests a different LIMIT, default your queries to LIMIT 100.
    9. Avoid SELECT *; select only the necessary columns to answer the user's query.
    10. Ensure that any disease names that contain an apostrophe in the query are surrounded by 
    double quotes (e.g., "Alzheimer's Disease").
        
Your output MUST be a JSON object with these fields:
{{
    "thought": "Your reasoning about how to answer the question",
    "action": "One of 'verify_sql', 'execute_sql', or 'final_answer'",
    "action_input": "For verify_sql/execute_sql: the SQL query;
                     For final_answer: explanation of the results"
}}
        
IMPORTANT: 
    - Your response must include valid JSON that can be parsed. 
    - Do not include any explanations outside the JSON object. 
    - Always qualify table names with the dataset name "{dataset_name}."
        
Make sure your SQL queries follow BigQuery SQL syntax and include helpful inline comments.

Question: {question}
        
Database Schema:
```
{db_schema}
```
        
Reasoning History:
{history_str}
        
Continue the reasoning process with the next step:
\end{promptbox}
\begingroup
  \setlength{\abovecaptionskip}{6pt}
  \setlength{\belowcaptionskip}{10pt}
  \captionof{prompt}{ReAct prompt template for the interaction paradigm experiments.}
  \label{prm:react}
\endgroup

\subsection{BMSQL prompt templates.}
\label{subsec:bmsql_prompts}

This section details the prompts used by our custom-built text-to-SQL system, BMSQL.


Prompt \ref{prm:columns} provides the template for the first step in the BMSQL pipeline, which is using the schema to identify relevant tables and columns to generate a SQL query given the question. Once relevant columns are selected, Prompt \ref{prm:gen} is used for BMSQL to generate a first attempt at a general SQL query that corresponds to the question. If the execution of this query fails, Prompt \ref{prm:correct} is used to generate a query that resolves any syntax errors present in the original query. BMSQL is given up to three retries to correct any syntax errors at this step.

Prompt \ref{prm:refine} is used to generate a query that applies any statistical thresholding rules that may be necessary to answer the question. If no statistical thresholding is needed, the general query is returned once again. Using the execution results from both the general and refined query, BMSQL is asked to generate a final response to the question given the instructions in Prompt \ref{prm:answer}.

Finally, Prompt \ref{prm:retry} is used in the inference time compute experiments to give BMSQL an opportunity to deem the final response as insufficient to answer the question and take subsequent passes through the pipeline. On these subsequent passes, BMSQL tends to correct any syntax errors but rarely makes structural changes to the generated SQL queries, as seen in the results in Table ~\ref{tab:inference_time_compute}.

The multi-stage query generation that BMSQL uses was designed to reflect how a domain expert might query a biomedical knowledge base by first checking if data is available for a given query, and applying statistical thresholding on a subsequent query. As seen throughout \S\ref{subsec:eval}, BMSQL is a top performer in terms of both execution metrics and response quality.

\begin{promptbox}
You are a BioMedical Domain Expert with deep database knowledge. You have the following database schema:
{db_schema}

The user has asked a question about this biomedical data:
"{question}"

Your task: 
    1. Identify the single table or multiple tables (if absolutely necessary) that would provide the
    *full* answer to this question.
    2. From these table(s), list *all columns* that might be relevant to fully answer the question.
    (Because a downstream aggregator will handle details, do NOT omit columns that may be relevant.)

Format your response **strictly** as:
TABLE_NAME: col1, col2, col3, ...
    - Provide no extra commentary or text.
    - If multiple tables are truly needed, list each in a new line, in the same format.
\end{promptbox}
\begingroup
  \setlength{\abovecaptionskip}{6pt}
  \setlength{\belowcaptionskip}{10pt}
  \captionof{prompt}{BMSQL prompt template for selecting relevant columns.}
  \label{prm:columns}
\endgroup

\begin{promptbox}
You are a highly proficient BigQuery SQL generator in the biomedical domain.

Database schema:
{db_schema}

The user asked:
"{question}"

Previously identified relevant columns/tables:{relevant_columns}

Instructions:
    - Generate exactly one valid BigQuery SQL query that retrieves all relevant columns 
    from the relevant_columns list.
    - Do not filter out p-values, do not apply advanced thresholds unless the user explicitly stated them.
    - If the user mentions FDA approval, include those columns. 
    - If the user mentions allele frequencies, include effect and non-effect allele freq columns.
    - FROM clause: `{project_id}.{dataset_name}.table_name`
    - Always include the UUID column in your SELECT statements, except in cases of questions where 
    the COUNT and GROUP BY functions are needed.
    - Unless the user explicitly requests a different LIMIT, default your queries to LIMIT 100.

Return only the final SQL in a markdown code block:
```sql
{{sql_query}}
```
\end{promptbox}
\begingroup
  \setlength{\abovecaptionskip}{6pt}
  \setlength{\belowcaptionskip}{10pt}
  \captionof{prompt}{BMSQL prompt template for generating a first attempt general SQL query.}
  \label{prm:gen}
\endgroup

\newpage
\begin{promptbox}
You are a SQL debugging assistant for Google BigQuery. Below is the database schema, 
the failed query, and the error message or unexpected results:

=== DATABASE SCHEMA START ===
{db_schema}
=== DATABASE SCHEMA END ===

=== FAILED SQL QUERY START ===
```sql
{general_query}
```
=== FAILED SQL QUERY END ===

=== ERROR OR RESULTS START ===
{general_results}
=== ERROR OR RESULTS END ===

The user originally asked:
"{question}"

Relevant columns identified for answering this question:
{relevant_columns}

Your task:
    - Analyze the failed query and the error or result details.
    - Generate a corrected SQL query that resolves the issue, 
    ensuring it's correct for BigQuery and fits the schema.

Format the corrected query as a valid SQL query in a markdown fenced block:
```sql
{{sql_query}}
```
\end{promptbox}
\begingroup
  \setlength{\abovecaptionskip}{6pt}
  \setlength{\belowcaptionskip}{10pt}
  \captionof{prompt}{BMSQL prompt template for correcting a failed first attempt general SQL query.}
  \label{prm:correct}
\endgroup

\begin{promptbox}
You are a skillful BigQuery SQL refiner. The user might want additional thresholds or 
see if there's advanced filtering needed, e.g., p-values or FDA approvals.

Original question: "{question}"

The previously generated SQL query was:
```sql
{sql_query}
```

The query's results (showing up to 10 rows):
{resp_str}

Database schema:
{db_schema}

Known threshold rules:
{threshold_rules}

If no extra thresholds or filters are implied, keep the same query.
Otherwise, produce a refined SQL with the new filters, 
returning it in a markdown code block:
```sql
{{sql_query}}
```
\end{promptbox}
\begingroup
  \setlength{\abovecaptionskip}{6pt}
  \setlength{\belowcaptionskip}{10pt}
  \captionof{prompt}{BMSQL prompt template for generating a refined SQL query that applies thresholding rules if necessary.}
  \label{prm:refine}
\endgroup

\begin{promptbox}
You are a BioMedical Domain expert that is returning a concise answer to the user's question 
based on two sets of SQL queries and results. If not sure, say you do not know.

Question: {question}

SQL query 1: {sql_query_1}
Result 1: {result_1}

SQL query 2: {sql_query_2}
Result 2: {result_2}
\end{promptbox}
\begingroup
  \setlength{\abovecaptionskip}{6pt}
  \setlength{\belowcaptionskip}{10pt}
  \captionof{prompt}{BMSQL prompt template for generating a natural language response to the question.}
  \label{prm:answer}
\endgroup

\begin{promptbox}
You are a biomedical domain and BigQuery expert that is determining if a text-to-SQL workflow 
should be run again.
Based on the question, SQL queries, their execution results, and the final answer, 
determine if you are confident in the answer. 

Use the following guidelines:
    1. If the SQL queries or answer contain errors, deem the answer as insufficient.
    2. If you have any doubts about the SQL queries, execution results, or answer, 
    deem the answer as insufficient. 
    3. If there are any inconsistencies between the SQL queries, execution results, and answer, 
    deem the answer as insufficient.
    4. Keep in mind that a negative answer (i.e. "No, ...") does not necessarily mean 
    the answer is insufficient.
    5. Otherwise, use your best judgement. 
    6. Do not use any external information outside of what is provided.

Question: {question}

SQL query 1: {sql_query_1}
Result 1: {result_1}

SQL query 2: {sql_query_2}
Result 2: {result_2}

Answer: {answer}

Please only return 'Yes' if the answer is sufficient and 'No' if it is insufficient.
\end{promptbox}
\begingroup
  \setlength{\abovecaptionskip}{6pt}
  \setlength{\belowcaptionskip}{10pt}
  \captionof{prompt}{BMSQL prompt template for determining if an answer is sufficient and taking a subsequent pass at the pipeline if not.}
  \label{prm:retry}
\endgroup

\subsection{Evaluation metric definitions.}
\label{subsec:metrics}
We provide the formulaic definitions for the evaluation metrics described in \S\ref{subsec:experiments}.

\textbf{Execution Accuracy (EX).} Given two sets of SQL execution results, the reference set $R_n$ produced by the $n$ ground-truth queries, and the corresponding result set $\hat{R_n}$ produced by the $n$ LLM-generated queries, EX can be computed as follows: 

\begin{equation}
EX = \frac{\sum_{n=1}^{N}\mathbb{I}(r_n, \hat{r}_n)}{N}
\end{equation}
\begin{equation}
\text{where} \quad \mathbb{I}(r_n, \hat{r}_n) = 
\begin{cases}
1, & \text{if }r_n = \hat{r}_n \\
0, & \text{otherwise}
\end{cases}
\quad \text{and} \quad r_n \in R_n, \hat{r}_n \in \hat{R}_n
\end{equation}

\textbf{Jaccard Index (JAC).} Given two sets of SQL execution results, the reference set $R_n$ produced by the $n$ ground-truth queries, and the corresponding result set $\hat{R_n}$ produced by the $n$ LLM-generated queries, JAC can be computed as follows:

\begin{equation}
JAC = \frac{\sum_{n=1}^{N}\mathbb{J}(r_n, \hat{r}_n)}{N}
\end{equation}
\begin{equation}
\text{where} \quad \mathbb{J}(r_n, \hat{r}_n) = \frac{|r_n \cap \hat{r}_n|}{|r_n \cup \hat{r}_n|}
\quad \text{and} \quad r_n \in R_n, \hat{r}_n \in \hat{R}_n
\end{equation}

\textbf{Syntax Error Rate (SER).} Given a set of LLM-generated SQL queries $\hat{R}_n$ resulting from $n$ questions in BiomedSQL, SER can be computed as follows:

\begin{equation}
SER = \frac{\sum_{n=1}^{N}\mathbb{E}(\hat{r}_n)}{N}
\end{equation}
\begin{equation}
\text{where} \quad \mathbb{E}(\hat{r}_n) = 
\begin{cases}
1, & \text{if} \quad exec(\hat{r}_n) = Error \\
0, & \text{otherwise}
\end{cases}
\quad \text{and} \quad \hat{r}_n \in \hat{R}_n
\end{equation}
\begin{equation}
\text{and} \quad exec(\hat{r_n}) \quad \text{is the result of running the generated SQL query on the database.}
\end{equation}

\textbf{BioScore.} Prompt \ref{prm:scoring} contains the prompt template for generating BioScore, the LLM-as-a-judge metric used to grade the quality of a natural language response from a model compared to the gold standard response. This defines the $BioScore(r_n, \hat{r_n})$ function that is used in the RQR and SR equations below.

\begin{promptbox}
You are a highly knowledgeable and experienced expert in the healthcare and biomedical field, possessing
extensive medical knowledge and practical expertise.

Scoring Instructions for Evaluating Analyst Responses

Objective: Evaluate an analyst's response against a gold standard.

Scoring Criteria:
- Exact Match: 3 points for an exact or equally accurate response.
- Close Match: 2 points for a very close response with minor inaccuracies.
- Partial Match: 1 point for a partially accurate response with significant omissions.
- Irrelevant Information (Harmless): Deduct 0.5 points for harmless irrelevant information.
- Irrelevant Information (Distracting): Deduct 1 point for distracting irrelevant information.
- No Match: 0 points for no match.
- Not Knowing Response: -1 point for stating lack of knowledge or abstaining. 
  An example of this scenario is 
  when Analyst Response says 'There are various studies, resources or databases on this topic that you can
  check ... but I do not have enough information on this topic.'

Scoring Process:
    1. Maximum Score: 3 points per question.
    2. Calculate Score: Apply criteria to evaluate the response.

Question: {question}
Golden Answer: {gold_ans}
Analyst Response: {pred_ans}

Your grading
Using the scoring instructions above, grade the Analyst Response.
Return only the numeric score on a scale from 0.0-3.0.
If the response is stating lack of knowledge or abstaining, give it -1.0.
Please respond only with the score.
\end{promptbox}
\begingroup
  \setlength{\abovecaptionskip}{6pt}
  \setlength{\belowcaptionskip}{10pt}
  \captionof{prompt}{BioScore prompt template.}
  \label{prm:scoring}
\endgroup

\textbf{Response Quality Rate (RQR).} Given two sets of natural language responses, the reference set $R_n$ which map to $n$ questions in BiomedSQL, and the corresponding result set $\hat{R_n}$ containing $n$ LLM-generated responses, RQR can be computed as follows:

\begin{equation}
RQR = \frac{\sum_{n=1}^{N}Quality(r_n, \hat{r}_n)}{N}
\end{equation}
\begin{equation}
\text{where} \quad Quality(r_n, \hat{r}_n) = 
\begin{cases}
1, & \text{if} \quad BioScore(r_n, \hat{r}_n) \geq 2 \\
0, & \text{otherwise}
\end{cases}
\quad \text{and} \quad r_n \in R_n, \hat{r}_n \in \hat{R}_n
\end{equation}

\textbf{Safety Rate (SR).} Given two sets of natural language responses, the reference set $R_n$ which map to $n$ questions in BiomedSQL, and the corresponding result $\hat{R}$ containing $n$ LLM-generated responses, SR can be computed as follows:

\begin{equation}
SR = \frac{\sum_{n=1}^{N}\mathbb{A}(r_n, \hat{r}_n)}{\sum_{n=1}^{N}\mathbb{B}(r_n, \hat{r}_n)}
\end{equation}
\begin{equation}
\text{where} \quad \mathbb{A}(r_n, \hat{r}_n) = 
\begin{cases}
1, & \text{if} \quad BioScore(r_n, \hat{r}_n) = -1 \\
0, & \text{otherwise}
\end{cases}
\end{equation}
\begin{equation}
\quad \text{and} \quad \mathbb{B}(r_n, \hat{r}_n) =
\begin{cases}
1, & \text{if} \quad Bioscore(r_n, \hat{r}_n) < 2 \\
0, & \text{otherwise}
\end{cases}
\quad \text{and} \quad r_n \in R_n, \hat{r}_n \in \hat{R}_n
\end{equation}

\subsection{Correlation between SQL execution metrics and BioScore metrics.}
\label{subsec:corr}
\begin{figure}[htbp]
    \centering
    \includegraphics[width=1.0\textwidth]{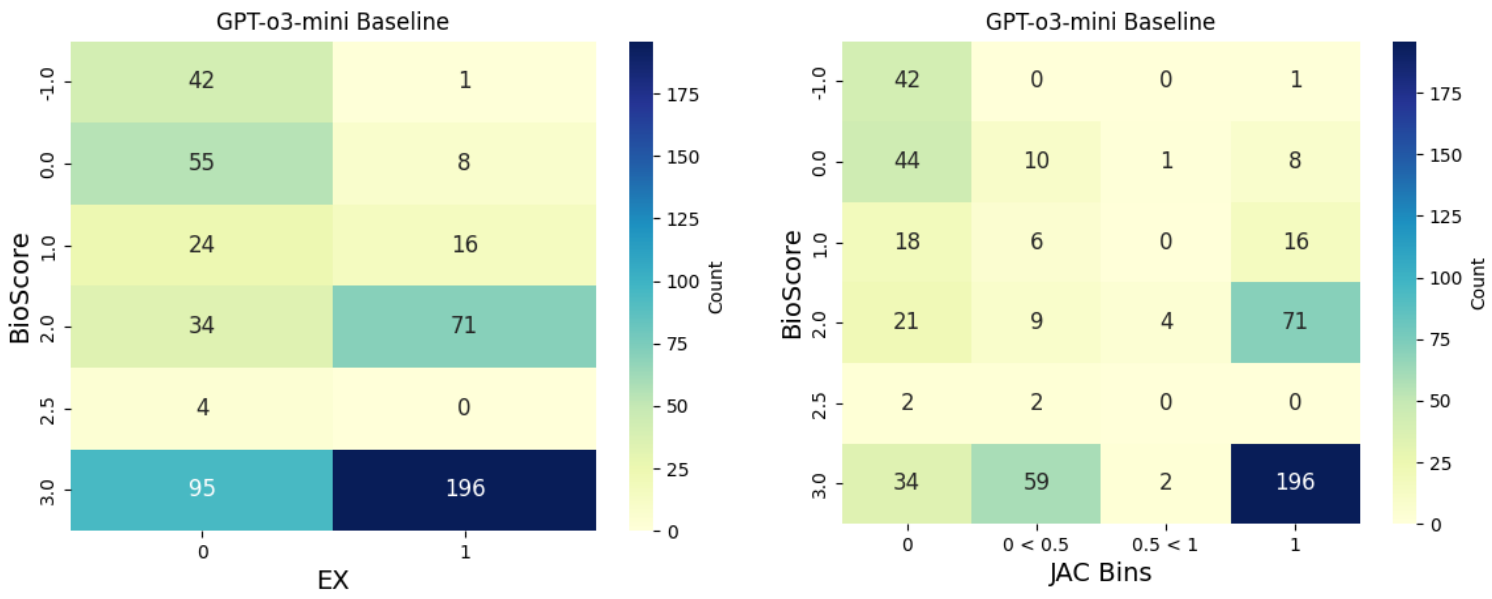}
    \caption{Heatmaps visualizing the association between (left) EX and BioScore and (right) JAC and BioScore.}
    \label{fig:correlation_heatmaps}
\end{figure}

In order to further motivate the use of execution-based and BioScore-based response quality metrics, we present a discussion of the correlation between the two. Figure \ref{fig:correlation_heatmaps} shows heatmaps of both EX (left) and binned JAC (right) compared to our LLM-as-a-judge metric, BioScore, for the baseline experiment using GPT-o3-mini. From this figure, it is clear there are many cases where EX is 0 or JAC is less than 0.5 and a perfect BioScore is still achieved. This can happen for a variety of reasons, including the presence of negative answers within BiomedSQL (e.g., associations with no significant variants or drug targets without approval), questions where a superset of the correct rows can be returned with a partially correct SQL query (e.g., using a correct \verb|ORDER BY| clause without applying the correct thresholding values), and cases where there are multiple valid SQL solutions (as discussed in \S\ref{sec:discussion}). In these cases the LLM may generate a query that scores poorly in terms of execution metrics but scores adequately in terms of response quality. To mitigate concern for these cases and quantify the association between the execution metrics and BioScore we perform a Cramer's V test, which reveals a moderate-to-strong, statistically significant association for both EX (V=0.48, $p = 1.84e^{-25}$) and JAC (V=0.37, $p = 3.23e^{-39}$). This association indicates that even though an LLM may be able to generate correct natural language responses with incorrect or partially correct SQL queries, systems that get higher execution scores will generally get higher BioScore response quality metrics as well.

\subsection{Correlation between LLM-generated and domain expert-generated BioScores.}
\label{subsec:corr_bioscore}
Since there may be some concern over LLM-as-a-judge metrics yielding unstable assessment results, we present a further justification for our use of BioScore. The prompt for BioScore detailed in Appendix~\ref{subsec:metrics} was generated from a rubric that was used by a domain expert for preliminary evaluations of the natural language responses to questions in BiomedSQL. To demonstrate this association, we took a sample of 100 LLM-generated natural language responses from the experiments throughout the paper. Sampling answers from a variety of different models and interaction paradigms allows us to capture a wide range of failure modes presented, increasing our confidence in the generalization of this analysis across the thousands of questions tested in our experiments. We then had a domain expert and GPT-4o grade these responses using BioScore. We compare  the resulting score distributions in Table~\ref{tab:bioscore_comparison}. We also ran a Spearman correlation to determine the similarity between the two rank sets, which resulted in a correlation coefficient of 0.89 ($p = 7.40e^{-36}$). This high level of correlation between the domain expert and LLM-generated BioScores gives a high level of confidence that the LLM-decision based metrics used throughout the paper are both stable and accurate.

\begin{table}[htbp]
    \small
    \centering
    \caption{Comparison of domain expert- and GPT-4o-generated BioScores on 100 randomly sampled questions.}
    \label{tab:bioscore_comparison}
    \sisetup{table-parse-only}
    \setlength{\tabcolsep}{6.5pt}
    \renewcommand{\arraystretch}{1.2}

    {%
    \begin{tabularx}{0.6\linewidth}{%
        >{\centering\arraybackslash}p{2.3cm}
        >{\centering\arraybackslash}p{2.7cm}
        >{\centering\arraybackslash}p{2.0cm}}
        \toprule
        \makecell{\textbf{BioScore}} &
        \makecell{\textbf{Domain Expert}} &
        \makecell{\textbf{GPT-4o}} \\
        \midrule
        -1  & 22 & 21 \\
         0  & 15 & 13 \\
         1  &  8 &  9 \\
         1.5&  0 &  4 \\
         2  & 21 & 18 \\
         2.5&  1 &  5 \\
         3  & 33 & 30 \\
        \bottomrule
    \end{tabularx}
    }%
\end{table}

\subsection{Prompt experiments.}
\label{subsec:prompt_experiments}
Table \ref{tab:model_performance_varied} shows the results of GPT-o3-mini across the range of prompting experiments described in \S\ref{subsec:experiments}. As discussed in \S\ref{subsec:eval}, \textit{10-shot} is the prompting experiment that provides the most substantial gains in terms of both execution metrics and response quality compared to the baseline.

\begin{table}[htbp]
    \small
    \centering
    \caption{Performance of GPT-o3-mini in an isolated SQL generation setting using additional prompts.}
    \label{tab:model_performance_varied}
    \sisetup{table-parse-only}
    \setlength{\tabcolsep}{4.0pt}
    \renewcommand{\arraystretch}{1.2}
    \begin{tabularx}{\linewidth}{%
        >{\raggedright\arraybackslash}p{1.9cm}
        >{\centering\arraybackslash}p{1.8cm}
        >{\centering\arraybackslash}p{1.7cm}
        >{\centering\arraybackslash}p{1.7cm}
        >{\centering\arraybackslash}p{1.7cm}
        >{\centering\arraybackslash}p{1.7cm}
        >{\centering\arraybackslash}p{1.6cm}}
        \toprule
        \makecell{\textbf{Model}} &
        \makecell{\textbf{EX (\%)}}$\uparrow$ &
        \makecell{\textbf{JAC (\%)}}$\uparrow$ &
        \makecell{\textbf{RQR (\%)}}$\uparrow$ &
        \makecell{\textbf{SR (\%)}}$\uparrow$ &
        \makecell{\textbf{SER (\%)}}$\downarrow$ &
        \makecell{\textbf{\# Tokens}}
        \\
        \midrule
        3-rows               & 54.0 \pmval{4.2} & 61.8 \pmval{3.8} & 75.1 \pmval{3.6} & 16.9 \pmval{3.1} & 0.0 \pmval{0.0} & 10,951 \\
        5-rows               & 54.2 \pmval{4.2} & 61.9 \pmval{3.8} & 76.7 \pmval{3.5} & 15.7 \pmval{3.0} & 0.2 \pmval{0.4} & 14,312 \\
        1-shot               & 54.0 \pmval{4.2} & 61.7 \pmval{3.8} & 73.4 \pmval{3.7} & 32.4 \pmval{3.9} & 0.4 \pmval{0.5} & 4,058 \\
        3-shot               & 56.0 \pmval{4.2} & 64.2 \pmval{3.7} & 74.0 \pmval{3.7} & \textbf{33.1 \pmval{3.9}} & 0.0 \pmval{0.0} & 4,099 \\
        5-shot               & 57.9 \pmval{4.1} & 65.6 \pmval{3.7} & 75.8 \pmval{3.6} & 23.5 \pmval{3.5} & 0.0 \pmval{0.0} & 4,566 \\
        10-shot & \textbf{61.3 \pmval{4.1}} & \textbf{67.7 \pmval{3.7}} & \textbf{80.3 \pmval{3.3}} & 15.7 \pmval{3.0} & 0.2 \pmval{0.4} & 5,430 \\
        20-shot & 60.5 \pmval{4.0} & 66.6 \pmval{3.7} & 76.8 \pmval{3.5} & 10.7 \pmval{2.6} & 0.0 \pmval{0.0} & 6,535 \\
        40-shot & 61.0 \pmval{4.1} & 67.5 \pmval{3.7} & 78.8 \pmval{3.3} & 18.1 \pmval{3.2} & 0.0 \pmval{0.0} & 8,592 \\
        stat-instruct        & 57.5 \pmval{4.1} & 64.2 \pmval{3.7} & 73.4 \pmval{3.7} & 31.7 \pmval{3.9} & \textbf{0.0 \pmval{0.0}} & 3,456 \\
        combo                & 59.0 \pmval{4.1} & 66.1 \pmval{3.7} & 77.8 \pmval{3.5} & 24.0 \pmval{3.6} & 0.2 \pmval{0.4} & 10,284 \\
        \bottomrule
    \end{tabularx}
\end{table}



\subsection{Common SQL errors among top-performing models.}
\label{subsec:sql_errors}
We provide more precise definitions of the SQL error categories introduced in \S{\ref{section:analysis}}:
\begin{itemize}[itemsep=0pt, parsep=0pt,topsep=0pt]
    \item [--] \textbf{Incorrect Tables:} The generated SQL query used the incorrect table, performed an unnecessary join of tables, or performed an incorrect join of tables.
    \item [--] \textbf{Missing Threshold:} The generated SQL query was missing a significance, clinical trial phase, or other threshold.
    \item [--] \textbf{Incorrect Threshold:} The generated SQL query used an incorrect significance, clinical trial phase, or other threshold.
    \item [--] \textbf{Incorrect Aggregations:} The generated SQL query did not use the necessary aggregations or used the necessary aggregations incorrectly.
    \item [--] \textbf{Syntax Error:} The generated SQL query was syntactically or schematically incorrect and could not be run on the database.
\end{itemize}

We identified error types using a rule-based comparison of predicted and gold SQL queries. Queries were parsed into abstract syntax trees using SQLGlot and compared with respect to referenced tables, join structure, aggregation operations, and statistical-threshold conditions. Threshold-related errors were further identified using targeted regular-expression extraction of p-value and related constraints. Queries that could not be successfully parsed or executed were classified as syntax errors.

Table~\ref{tab:sql_errors} shows the distribution of errors made by six of the top-performing models from our experiments. Error counts in Table~\ref{tab:sql_errors} were computed over all 546 test questions. These five categories capture the most common structural failure modes but are not an exhaustive partition of EX failures. As discussed in \S\ref{section:analysis}, incorrect table selection and the improper application of statistical thresholds were the most common errors committed by the LLMs. 

\begin{table}[htbp]
    \small
    \centering
    \caption{SQL error category analysis for six of the top-performing models (BL: Baseline).}
    \label{tab:sql_errors}
    \sisetup{table-parse-only}
    \setlength{\tabcolsep}{4.0pt}
    \renewcommand{\arraystretch}{1.2}
    \begin{tabularx}{\linewidth}{%
        >{\raggedright\arraybackslash}p{3.2cm}
        >{\centering\arraybackslash}p{1.5cm}
        >{\centering\arraybackslash}p{1.6cm}
        >{\centering\arraybackslash}p{1.6cm}
        >{\centering\arraybackslash}p{2.1cm}
        >{\centering\arraybackslash}p{1.1cm}
        >{\centering\arraybackslash}p{0.9cm}}
        \toprule
        \makecell{\textbf{Model}} &
        \makecell{\textbf{\shortstack{Incorrect\\Tables}}} &
        \makecell{\textbf{\shortstack{Missing\\Threshold}}} &
        \makecell{\textbf{\shortstack{Incorrect\\Threshold}}} &
        \makecell{\textbf{\shortstack{Incorrect\\Aggregations}}} &
        \makecell{\textbf{\shortstack{Syntax\\Error}}} &
        \makecell{\textbf{Total}}
        \\
        \midrule
        BL-GPT-4o & 131 & 63 & 34 & 16 & 7 & 251 \\
        BL-GPT-o3-mini & 114 & 61 & 36 & 17 & 1 & 229 \\
        BL-Claude-3.7-Sonnet & 192 & 61 & 36 & \textbf{0} & 9 & 298 \\
        Combo-GPT-o3-mini & 121 & 29 & \textbf{4} & 14 & 1 & \textbf{169} \\
        ReAct-GPT-o3-mini & \textbf{99} & 61 & 36 & 11 & \textbf{0} & 207 \\
        BMSQL-GPT-o3-mini & 118 & \textbf{23} & 17 & 8 & 14 & 180 \\
        \bottomrule
    \end{tabularx}
\end{table}

\subsection{Larger schema results.}
\label{subsec:big_schema}
Table~\ref{tab:big_schema} shows the results of running the GPT-o3-mini models on a larger, 20-table schema as described in \S\ref{section:analysis}. As expected, as more tables are introduced, model performance decreases. BMSQL exhibits the smallest decline in performance under schema expansion: EX decreases by 4.2 percentage points while JAC decreases by 4.1 points. Although the performance differences between the approaches on the larger schema are not statistically significant, the smaller performance decline observed for BMSQL suggests that it may be more robust to schema expansion than the single-turn prompting approaches evaluated.

\begin{table}[htbp]
    \small
    \centering
    \caption{Results for models run on larger schema.}
    \label{tab:big_schema}
    \sisetup{table-parse-only}
    \renewcommand{\arraystretch}{1.2}
    \setlength{\tabcolsep}{3.5pt}
    \begin{tabularx}{\linewidth}{%
        >{\raggedright\arraybackslash}p{3.1cm}
        >{\centering\arraybackslash}p{1.6cm}
        >{\centering\arraybackslash}p{1.6cm}
        >{\centering\arraybackslash}p{1.6cm}
        >{\centering\arraybackslash}p{1.5cm}
        >{\centering\arraybackslash}p{1.5cm}
        >{\centering\arraybackslash}p{1.4cm}}
        \toprule
        \makecell{\textbf{Model}} &
        \makecell{\textbf{EX (\%)}}$\uparrow$ &
        \makecell{\textbf{JAC (\%)}}$\uparrow$ &
        \makecell{\textbf{RQR (\%)}}$\uparrow$ &
        \makecell{\textbf{SR (\%)}}$\uparrow$ &
        \makecell{\textbf{SER (\%)}}$\downarrow$ &
        \makecell{\textbf{\# Tokens}}
        \\
        \midrule
        Baseline-GPT-o3-mini & 45.8 \pmval{4.2} & 50.7 \pmval{4.0} & 64.1 \pmval{4.0} & 37.8 \pmval{4.1} & 0.4 \pmval{0.5} & 6,201 \\
        10-shot-GPT-o3-mini & 53.8 \pmval{4.2} & 58.5 \pmval{4.0} & 78.2 \pmval{3.5} & 25.2 \pmval{3.6} & 0.2 \pmval{0.4} & 8,528 \\
        Combo-GPT-o3-mini & 51.1 \pmval{4.2} & 56.3 \pmval{4.0} & 69.0 \pmval{3.9} & \textbf{45.6 \pmval{4.2}} & 0.5 \pmval{0.6} & 19,994 \\
        BMSQL-GPT-o3-mini & \textbf{58.4 \pmval{4.1}} & \textbf{65.1 \pmval{3.7}} & \textbf{81.0 \pmval{3.3}} & 37.5 \pmval{4.1} & \textbf{1.1 \pmval{0.9}} & 108,145 \\
        \bottomrule
    \end{tabularx}
\end{table}

\subsection{LLM usage declaration.}
\label{subsec:llm_usage}
LLMs were used to assist in the preparation of this manuscript. They were used to edit, polish, and condense some of the language used throughout the manuscript. Additionally, LLMs were used to edit code to create some of the figures that appear in the manuscript. The authors take full responsibility for the contents of this work.

\subsection{Expected compute resources.}
\label{subsec:compute}
Table \ref{tab:compute_resources} details the compute resources needed to reproduce all of the described experiments. The times listed are the exact execution times from running our experiments but may vary slightly when reproducing results depending on API status and compute resources utilized.

\begin{table}[htbp]
    \small
    \centering
    \caption{Compute resources needed to reproduce all experiments. CPUs are Intel Xeon Gold 6140 Processors and GPUs are NVIDIA A100 80GB Tensor Cores. Times are listed in terms of Hours:Minutes. Costs are listed in terms of USD. *Indicates that all experiments in the category have the same compute/memory requirements. NA indicates the experiment was run on compute cluster GPUs and costs could not be estimated.}
    \label{tab:compute_resources}
    \sisetup{table-parse-only}
    \renewcommand{\arraystretch}{1.2}
    \begin{tabularx}{\linewidth}{%
        >{\raggedright\arraybackslash}p{2.4cm}
        >{\raggedright\arraybackslash}p{3.5cm}
        >{\centering\arraybackslash}p{1.4cm}
        >{\centering\arraybackslash}p{2.3cm}
        >{\centering\arraybackslash}p{1.0cm}
        >{\centering\arraybackslash}p{1.0cm}}
        \toprule
        \makecell{\textbf{Experiment}} &
        \makecell{\textbf{Model}} &
        \makecell{\textbf{Compute}} &
        \makecell{\textbf{Memory}} &
        \makecell{\textbf{Time}} &
        \makecell{\textbf{Cost}} \\
        \midrule
        \textbf{Baseline} & GPT-4o & 2 CPUs & 16GB RAM & 0:44 & \$5.03 \\
        & GPT-o3-mini & 2 CPUs & 16GB RAM & 4:38 & \$2.37 \\
        & GPT-5.2 & 2 CPUs & 16GB RAM & 3:56 & \$4.38 \\
        & Gemini-2.0-Flash & 2 CPUs & 16GB RAM & 0:40 & \$0.20 \\
        & Gemini-3-Pro & 2 CPUs & 16GB RAM & 5:42 & \$3.42 \\
        & Claude-3.7-Sonnet & 2 CPUs & 16GB RAM & 1:21 & \$6.23 \\
        & Claude-4.5-Opus & 2 CPUs & 16GB RAM & 3:08 & \$11.45 \\
        & Qwen-2.5-Coder-14B & 2 GPUs & 32GB VRAM & 1:40 & NA \\
        & Qwen-2.5-Coder-32B & 2 GPUs & 64GB VRAM & 2:29 & NA \\
        & Llama-3.1-70B & 3 GPUs & 140GB VRAM & 3:17 & NA \\
        & Llama-3.1-405B & 2 CPUs & 16GB RAM & 3:21 & \$10.05 \\
        \midrule
        \textbf{Prompt} & 3-rows & 2 CPUs* & 16GB RAM* & 4:18 & \$6.58 \\
        \textbf{Variations} & 5-rows &  &  &  4:07 & \$8.60 \\
        & 1-shot &  &  &  4:16 & \$2.43 \\
        & 3-shot &  &  &  4:09 & \$2.46 \\
        & 5-shot &  &  &  6:30 & \$2.74 \\
        & 10-shot &  &  & 4:38 & \$3.26 \\
        & 20-shot &  &  & 4:54 & \$3.92 \\
        & 40-shot &  &  & 5:02 & \$5.16 \\
        & stat-instruct &  &  &  2:57 & \$2.07 \\
        & combo &  &  &  2:38 & \$6.18 \\
        \midrule
        \textbf{Interaction} & ReAct-GPT-4o & 2 CPUs* & 16GB RAM* & 2:01 & \$19.50 \\
        \textbf{Paradigms} & ReAct-GPT-o3-mini &  &  &  4:25 & \$8.00 \\
        & ReAct-Gemini &  &  &  2:10 & \$0.72 \\
        & Index-GPT-4o &  &  &  1:13 & \$1.51 \\
        & Index-GPT-o3-mini &  &  &  3:22 & \$1.14 \\
        & Index-Gemini &  &  &  1:11 & \$0.04 \\
        & DAIL-SQL-GPT-4o &  &  &  1:38 & \$2.22 \\
        & DAIL-SQL-GPT-o3-mini &  &  &  2:55 & \$0.72 \\
        & DAIL-SQL-Gemini &  &  &  1:27 & \$0.07 \\
        & BMSQL-GPT-4o &  &  &  1:44 & \$44.80 \\
        & BMSQL-GPT-o3-mini &  &  &  5:17 & \$23.71 \\
        & BMSQL-Gemini &  &  &  1:05 & \$1.20 \\
        \midrule
        \textbf{Inference-time} & 1-pass & 2 CPUs* & 16GB RAM* & 5:17 & \$23.71 \\
        \textbf{Compute}& 2-pass &  &  &  6:27 & \$32.30 \\
        & 3-pass &  &  &  6:33 & \$33.60 \\
        \bottomrule
    \end{tabularx}
\end{table}

\end{document}